\documentclass[letterpaper, 10 pt, conference]{ieeeconf}  % Comment this line out if you need a4paper

\IEEEoverridecommandlockouts                              % This command is only needed if 
\title{\LARGE \bf
Disentangled Point Diffusion for Precise Object Placement
}

\author{Lyuxing He*$^{1}$, Eric Cai*$^{1}$, Shobhit Aggarwal$^{1}$, Jianjun Wang$^{2}$, David Held$^{1}$%
\thanks{* Equal contribution}
\thanks{$^{1}$ The authors are with Carnegie Mellon University, Pittsburgh, USA.
         {\tt\scriptsize \{lyuxingh, eycai, shobhita, dheld\}@andrew.cmu.edu}}%
\thanks{$^{2}$ The author is with ABB Inc., USA.
         {\tt\scriptsize jianjun.wang@us.abb.com}}%
\thanks{David Held holds concurrent appointments at CMU and as an Amazon Scholar. This paper describes work performed at CMU and is not associated with Amazon.}
}

\usepackage{url}
\usepackage{comment}
\usepackage{xcolor}
\usepackage{amsmath}
\usepackage{amssymb}
\usepackage{booktabs}    % for \toprule, \midrule, \bottomrule
\usepackage{array}       % for table column alignment options
\usepackage{wrapfig}     % for wraptable environment
\usepackage{graphicx}
\usepackage[caption=false,font=footnotesize]{subfig}
\usepackage{mathtools}
\usepackage{cuted}
\usepackage{caption} % put this in your preamble
\usepackage{xr}   % or \usepackage{xr} if you don’t need hyperlinks
\usepackage{stfloats}
\usepackage{multirow}
\usepackage{makecell}

\definecolor{lightblue}{rgb}{0.42, 0.81, 0.89}  % light blue RGB

\definecolor{turquoise}{rgb}{0.25, 0.88, 0.82}  % light blue RGB

\begin{document}

\maketitle
\thispagestyle{empty}
\pagestyle{empty}

\begin{strip}
\vspace{-3.5\baselineskip} % pull up the figure
\centering
\includegraphics[width=\linewidth]{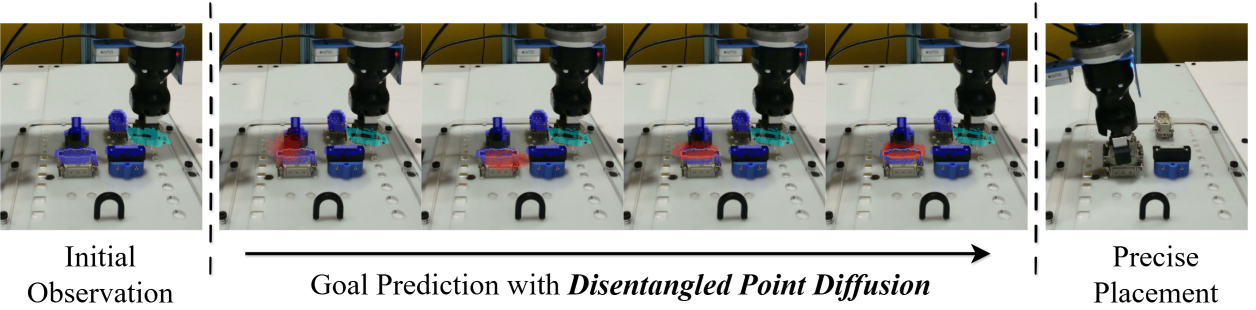}
\captionof{figure}{Our method (TAX-DPD) uses \emph{disentangled point diffusion} to predict precise goal configurations for a millimeter-precision industrial insertion task. \textcolor{blue}{Blue} denotes the scene point cloud, \textcolor{turquoise}{turquoise} denotes the manipulated object point cloud, and \textcolor{red}{red} denotes the diffused goal point cloud, where we jointly diffuse the object placement frame and geometry.}

\label{fig:pull}
\vspace{-1\baselineskip}
\end{strip}

%%%%%%%%%%%%%%%%%%%%%%%%%%%%%%%%%%%%%%%%%%%%%%%%%%%%%%%%%%%%%%%%%%%%%%%%%%%%%%%%
%\vspace{20pt}
\begin{abstract}

Recent advances in robotic manipulation have highlighted the effectiveness of learning from demonstration. However, while end-to-end policies excel in expressivity and flexibility, they struggle both in generalizing to novel object geometries and in attaining a high degree of precision. An alternative, object-centric approach frames the task as predicting the placement pose of the target object, providing a modular decomposition of the problem. Building on this goal-prediction paradigm, we propose TAX-DPD, a hierarchical, disentangled point diffusion framework that achieves state-of-the-art performance in placement precision, multi-modal coverage, and generalization to variations in object geometries and scene configurations. We model global scene-level placements through a novel feed-forward Dense Gaussian Mixture Model (GMM) that yields a spatially dense prior over global placements; we then model the local object-level configuration through a novel disentangled point cloud diffusion module that separately diffuses the object geometry and the placement frame, enabling precise local geometric reasoning.  Interestingly, we demonstrate that our point cloud diffusion achieves substantially higher accuracy than a prior approach based on SE(3)-diffusion, even in the context of rigid object placement. We validate our approach across a suite of challenging tasks in simulation and in the real-world on high-precision industrial insertion tasks. Furthermore, we present results on a cloth-hanging task in simulation, indicating that our framework can further relax assumptions on object rigidity. Visualizations and supplementary materials can be found on our project website: \url{https://3dgp-icra2026.github.io/}.

\end{abstract}

%%%%%%%%%%%%%%%%%%%%%%%%%%%%%%%%%%%%%%%%%%%%%%%%%%%%%%%%%%%%%%%%%%%%%%%%%%%%%%%%
\section{INTRODUCTION}
Learning from demonstration has emerged as a popular paradigm for robotic manipulation~\cite{zhaolearning, zhaoaloha2, wutidybot++, chi2024universal, wu2024gello, chengopen, chi2023diffusion, ze2024dp3, shafiullah2022behavior, lee2024behavior}. 
%} and expressive neural policy architectures~\cite{
While direct end-to-end visuomotor policy learning methods
%, which jointly learn \emph{where} to place and \emph{how} to act, 
have produced impressive results across a diverse range of complex manipulation tasks, 
(e.g. shoelace tying~\cite{zhaoaloha2}, sauce pouring~\cite{chi2023diffusion}, laundry folding~\cite{black2024pi_0}), they have yet to display robust generalization to variations in object geometry, nor have they shown the precision required for low-tolerance tasks, such as industrial manufacturing or inserting a key into a lock.  %object- and scene-level variations requisite for broad, in-the-wild deployment.

An alternative line of work is to modularly decompose the problem into \emph{where} and \emph{how}—first predicting a goal configuration (\emph{where to place an object}) and then executing it with a low-level policy or motion planner (\emph{how to place the object at that location}).   Such a decomposition enables the system to reason more explicitly about object-centric geometry as a form of goal prediction~\cite{cai2024tax3d, pan2023tax, huang2024imagination, simeonov2023shelving, zhao2025anyplace, liu2023structdiffusion, wang2024learning, eisnerdeep, simeonov2022neural, florence2018dense, chang2024dap}. Recent methods further leveraged generative approaches to capture multi-modal placement distributions~\cite{simeonov2023shelving, zhao2025anyplace, liu2023structdiffusion, wang2024learning, chang2024dap}, providing multiple feasible solutions for tasks where diverse goal configurations are possible. Although these works show a high degree of sample efficiency and some degree of generalization to novel object geometries, they still do not achieve the precision required for very low-tolerance tasks.

We posit that this limitation arises from the dominant reliance on SE(3)-based representations for goal prediction. Although an SE(3) pose is sufficient to represent the configuration of a single object, it is difficult to define a consistent SE(3) pose representation across a range of objects with varying geometries. An alternative approach is to predict the goal configuration via point cloud generation~\cite{huang2024imagination,cai2024tax3d}, i.e. predicting the point cloud of the target object in the goal configuration. Point cloud generation avoids the need to define a consistent reference frame across a range of objects with varying geometries. From the generated point cloud, one can derive the SE(3) transformation to perform the precise placement.  \textit{We show that point cloud generation leads to significantly more accurate object placement, even when placing a rigid object, when generalizing across a class of objects of varying geometry.}

%which can learn an explicit, frame-agnostic geometric state. Nonetheless, its direct application to object manipulation is problematic. Specifically, the standard practice of scale normalization essential for stable training in generative models is fundamentally at odds with the demands of precise relative placement. In such tasks, goal configurations can span an entire scene, creating an inherent conflict between preserving an object's fine-grained local geometry and its absolute global position within that larger workspace. More fundamentally, existing generative approaches, whether diffusing poses or points, introduce an inherent trade-off between coverage of diverse placements and the variance of any single prediction, undermining the precision required for successful task completion.

%Our insight is that we can achieve higher precision by diffusing object point clouds rather than diffusing SE(3) poses, as is done in prior work (cite RPDiff and AnyPlace).  We show that point cloud diffusion leads to much more accurate placements, even when evaluated on rigid objects for which SE(3) poses are sufficient, when trained over a variations in object shape and geometry.  Our insight is that point cloud diffusion, in contrast to SE(3) diffusion, enables our method to reason about low-level details of the object geometry, enabling precise object insertion. \lyuxing{might need some rephrasing here} 

However, we found that previous approaches to point cloud diffusion struggle to produce high-fidelity configurations when modeling multi-modal placements across large scenes. %, limiting the precision that is required for reliable object placement. 
Our second insight is that explicitly decoupling point cloud diffusion into distinct problems of multi-modal coverage, shape prediction, and frame prediction can overcome these challenges. To materialize these insights, we propose \textbf{TAX-DPD} (\textbf{TA}sk-specific \textbf{Cross}-Geometry reasoning with \textbf{D}isentangled \textbf{P}oint \textbf{D}iffusion), a hierarchical framework that operates in two stages: 1) \textbf{global placement initialization}, where a neural network is trained to predict a \emph{Dense Gaussian Mixture Model (GMM)} to capture the multi-modal distribution of potential object placement positions  %frame %local placement frames 
across the scene, ensuring coverage over placement locations; and 2) \textbf{local configuration refinement}, where a novel \emph{disentangled point diffusion} process predicts the  object placement configuration, by separately denoising the object-centric geometry and object frame in the initialized placement frame. For rigid objects, we further recover the final SE(3) pose using a standard RANSAC-SVD alignment procedure. % Through jointly denoising the object's configuration and its extrinsic placement frame, we thereby unify the robust geometric representation of point diffusion with the high-fidelity placement of SE(3) diffusion. 
Concretely, our contributions are as follows:
\begin{enumerate}
\item A novel global placement initialization method using a deep network to predict a Dense GMM to model multi-modal placement distributions at the scene-level.
\item A novel local configuration refinement method using a disentangled point diffusion objective for the separate denoising of object geometry and placement frame, allowing for precise placement predictions.
\item A broad suite of evaluations on simulation (mug-hanging, book-shelving, etc.) and real-world industrial insertion tasks, in which TAX-DPD achieves millimeter-level precision while also maintaining broad coverage for multi-modal tasks.
\end{enumerate}

\begin{comment}
Motivated by these challenges, we propose a hierarchical, generative system towards goal prediction for multi-modal and precise object placement. Our approach consists of two stages: 1) \textbf{global placement initialization}, in which we sample an approximate placement location within the larger scene through a learned, spatially-grounded \emph{Dense Gaussian Mixture Model (GMM)}, and 2) \textbf{local configuration refinement}, in which we predict the precise object–scene configuration via a joint diffusion process over object geometry and placement frame in point cloud space. Concretely, our contributions are as follows:
\begin{enumerate}
    \item A novel global placement initialization method using a Dense GMM to model multi-modal placement distributions at the scene-level.
    \item A novel local configuration refinement method using a disentangled point diffusion objective for the separate denoising of object geometry and placement frame, allowing for precise and dense per-point transformations.
    \item A broad suite of evaluations on simulation (mug-hanging, book-shelving, etc.) and real-world industrial insertion tasks, in which our method achieves state-of-the-art generalization and distributional coverage, while accomodating millimeter-level precision.
\end{enumerate}
\end{comment}

\section{RELATED WORK}
\subsection{Point Cloud Generation.}
Recent advancements in generative modeling have significantly enhanced the synthesis of 3D point clouds. Methods based on unconditional generative models including VAEs~\cite{brock2016generative, kim2021setvae}, GANs~\cite{achlioptas2018learning, shu20193d, yang2021cpcgan}, and Diffusion Models~\cite{luo2021diffusion, zyrianov2022learning, mo2023dit, mo2024fast, huang2022city3d} have demonstrated the ability to produce diverse and high-fidelity 3D shapes. Building upon these foundations, recent works explored conditioning point clouds generation on auxiliary information such as images \cite{dahnert2024coherent}, textual descriptions \cite{nichol2022point, sanghi2022clip}, partial point clouds \cite{lee2023diffusion}, or a diverse set of inputs \cite{ran2024towards, Zhou_2021_ICCV}, allowing for more controlled and context-aware synthesis. The success of these models stems from their ability to learn a rich, continuous latent representation of 3D shapes. This capacity for learning the underlying manifold of 3D data endows them with strong generalization capabilities, enabling them to synthesize novel object instances that structurally adhere to learned geometric priors. While prior works generate 3D shapes for data synthesis, TAX-DPD generates task-specific point clouds conditioned on a scene to infer relative transformations for object manipulation.

\subsection{Relative Placement Tasks.} Many placement tasks can be decomposed into predicting the geometric relationship between a pair of objects. Some prior work~\cite{simeonov2022neural, florence2018dense, pan2023tax, eisnerdeep} explicitly model this relationship by learning either category-level descriptors or dense object correspondences, from which a task-specific SE(3) transformation between objects (i.e. a goal pose) can be predicted and executed on a robot. To better accommodate multi-modal placements, some prior work~\cite{wang2024learning, simeonov2023shelving, chang2024dap, zhao2025anyplace} adopt generative methods to learn placement distributions defined directly in the SE(3) space, from which a goal pose can be sampled. Similar to TAX-DPD, another line of works leverages point cloud generation and denoises goal states through dense point flow~\cite{huang2024imagination} or point cloud diffusion~\cite{cai2024tax3d}. In contrast, we propose a hierarchical point cloud generation framework with a novel disentangled diffusion objective over object geometry and placement frames, enabling robust handling of high-precision and multi-modal relative placement tasks and strong generalization to novel object geometries and scene configurations.

\section{Problem Statement}

\subsection{Goal Prediction for Object Placement.} 
\begin{figure}[t]
    \centering
    \includegraphics[width=\linewidth]{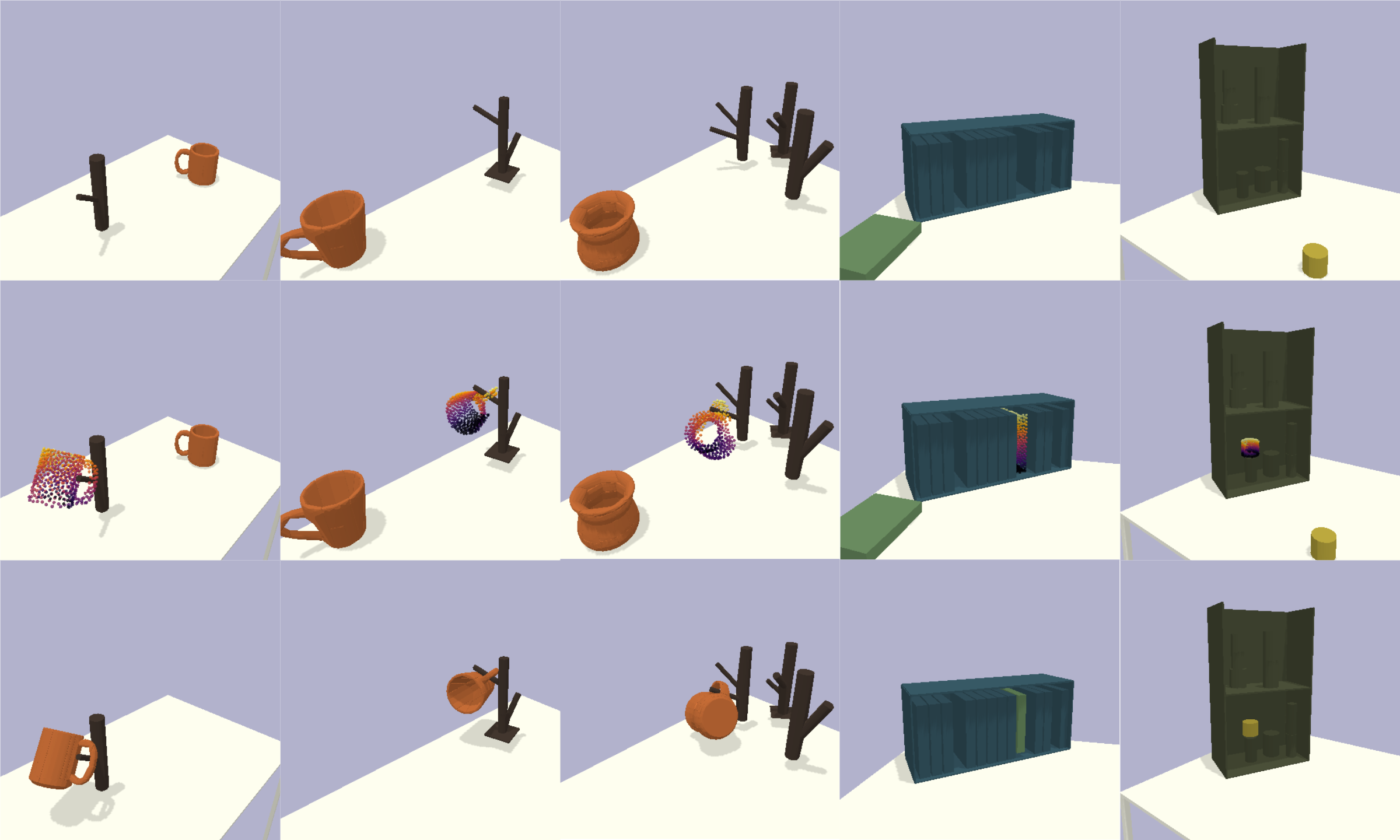}
    \caption{\textbf{RPDiff Task Environments.} (\textit{Top}) Our experiments span various multi-modal placement tasks with significant object and scene variation. (\textit{Middle}) TAX-DPD is able to precisely model goal configuration as point clouds. (\textit{Bottom}) Successful executions of our model's goal predictions.}
    \label{fig:environments}
\end{figure}

In this paper, we focus on general object placement tasks, in which an object $\mathcal{O}$ must be manipulated into a precise configuration within a larger scene $\mathcal{S}$ (e.g. hanging a mug on one of multiple racks - see Fig.~\ref{fig:environments}). %\eric{could ref fig here}). 
To solve this task, we predict the goal configuration of object $\mathcal{O}$ as a dense point cloud, i.e. the location that \textit{every point} in object $\mathcal{O}$ must move to successfully complete the task. More formally, given point clouds $P_\mathcal{O} \in \mathbb{R}^{N_{\mathcal{O}} \times 3}$ for the segmented object $\mathcal{O}$ and $P_\mathcal{S} \in \mathbb{R}^{N_{\mathcal{S}} \times 3}$ for the scene $\mathcal{S}$, we aim to predict a goal point cloud $\hat{P}^*_\mathcal{O} \in \mathbb{R}^{N_{\mathcal{O}} \times 3}$ such that $(\hat{P}^*_\mathcal{O}, P_\mathcal{S})$ represents a valid placement of $\mathcal{O}$. Since there are often multiple feasible placements for object $\mathcal{O}$, we %cast the prediction of the goal configuration $\hat{P}^*_\mathcal{O}$ as sampling 
aim to learn a distribution over goal point clouds that we can sample from, i.e. $\hat{P}^*_\mathcal{O} \sim f(P_\mathcal{O}, P_\mathcal{S})$. We then move object $\mathcal{O}$ to the goal configuration given by $\hat{P}^*_\mathcal{O}$ using motion planning or potentially a learned policy (see Appendix I on the project website for details).

This formulation stands in contrast to many existing methods for object-centric placement, which represent goals explicitly as SE(3) transformations under a rigid object assumption. Our experiments indicate that, by predicting point cloud configurations instead of SE(3) poses, TAX-DPD demonstrates greater precision when the task requires generalization across variations in object geometry, for which a unified pose reference frame can be difficult to define. Furthermore, we show that this point cloud–based formulation can be naturally relaxed to the setting of placing non-rigid objects. %, where we show that our method achieves superior performance to some limited initial work~\cite{cai2024tax3d}. 
%Additionally, there are often multiple feasible placements for object $\mathcal{O}$; to capture this multi-modality, we cast the prediction of goal configuration $\hat{P}^*_\mathcal{O}$ as the sampling from a learned distribution of placements given an object $\mathcal{O}$ and scene $\mathcal{S}$, i.e. $\hat{P}^*_\mathcal{O} \sim f(P_\mathcal{O}, P_\mathcal{S})$.

\subsection{Assumptions.} Similar to prior works on relative placement~\cite{simeonov2022neural, florence2018dense, pan2023tax, eisnerdeep}, we assume the object $\mathcal{O}$ that is being manipulated is segmented from the rest of the scene $\mathcal{S}$. During training, we assume access to a set of $N$ demonstrations $\{(P_{\mathcal{O}}^{(n)}, P_{\mathcal{S}}^{(n)}, {P^*}_{\mathcal{O}}^{(n)})\}_{n=1}^{N}$ which indicate the initial object point cloud $P_{\mathcal{O}}^{(n)}$, the initial scene point cloud $P_{\mathcal{S}}^{(n)}$, and the point cloud of the object in a goal configuration ${P^*}_{\mathcal{O}}^{(n)}$.  We further assume knowledge of the ground-truth correspondences between the initial object point cloud $P_\mathcal{O}^{(n)}$ and the goal point cloud $P_\mathcal{O}^{*(n)}$, which (for rigid objects) can be obtained from the demonstrations using the robot kinematics.

\begin{figure*}[t]
    \centering
    \includegraphics[width=\linewidth]{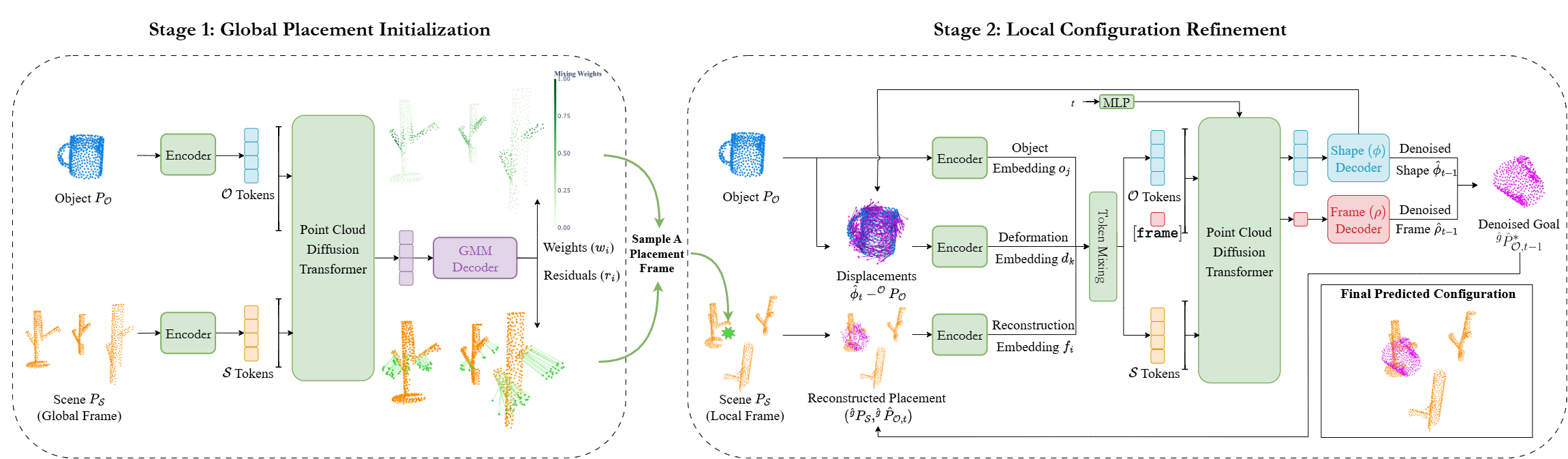}
    \caption{\textbf{Method Overview.} (\textit{Left}) Our \emph{Global Placement Initialization} samples a rough global position using a novel dense GMM-based prediction module, a framework that models highly multi-modal placement distributions at the scene-level. (\textit{Right}) Our \emph{Local Configuration Refinement} then proceeds with a novel disentangled shape and reference frame diffusion that simultaneously allow precise and dense goal predictions.}
    \label{fig:system-local}
\end{figure*}

\section{Method}
To sample a goal configuration $\hat{P}^*_\mathcal{O} \sim f(P_\mathcal{O}, P_\mathcal{S})$, our approach builds on Denoising Diffusion Probabilistic Models (DDPM)~\cite{ho2020denoising, nichol2021improved}, in which data samples $x_0$ are perturbed under a Markovian noising process $q(x_t | x_{t - 1}) = \mathcal{N}(x_t|\sqrt{1 - \beta_t}x_{t - 1}, \beta_tI)$ and a network is trained to learn the reverse transitions $p_{\theta}(x_{t-1}|x_t) = \mathcal{N}(x_{t-1}|\mu_{\theta}(x_t, t), \sigma^2_tI)$. While Denoising Diffusion Probabilistic Models (DDPMs) have demonstrated a powerful capacity to learn rich geometric distributions in the point cloud generation domain~\cite{luo2021diffusion, nichol2022point, Zhou_2021_ICCV, mo2023dit, mo2024fast}, we find that prior works integrating them for goal-prediction in object manipulation~\cite{cai2024tax3d, huang2024imagination, zhen20243d} struggle with obtaining high precision.  %performance deterioration due to the large scale discrepancy in scale between the object and its surrounding scene, which is further exacerbated in multi-modal tasks where placement modes are spanning across the workspace. 
To leverage the powerful capacity of point cloud diffusion for the distinct challenge of simultaneously achieving high precision, multi-modal coverage, and robustness to unseen object geometries, %we propose several key methodological advancements, detailed in the following sections:
we propose the following two key methodological contributions (1 and 2), complemented by a rigid alignment procedure (3):
\begin{enumerate}
    \item \textit{Global placement initialization} (Fig.~\ref{fig:system-local}, \textit{left}):  We sample an approximate placement frame $\hat{g}\in\mathbb{R}^3$ (i.e. the  centroid of the object in the goal pose) from a dense Gaussian Mixture Model (GMM) that is predicted per scene by a network $\hat{g}\sim f_{\text{global}}(P_\mathcal{O},P_\mathcal{S})$. 
    \item \textit{Local placement refinement} (Fig.~\ref{fig:system-local}, \textit{right}):  We perform a disentangled point cloud diffusion in the reference frame $\hat{g}$. We decompose the prediction of the goal $^{\hat{g}}P^*_{\mathcal{O}}$ in the local placement frame $\hat{g}$ into a mean-centered shape $\phi$ and an object frame $\rho$ prediction, with $^{\hat{g}}P^*_{\mathcal{O}}=\phi+\rho$ and $\hat{P}^*_\mathcal{O}={^{\hat{g}}\hat{P}^*_{\mathcal{O}}}+\hat{g}$ (detailed definitions in Sec.~\ref{sec:Local Placement Refinement}). Following standard DDPM practice, we gradually denoise these variables via an iterative process: $f_{\text{local}}\!\left(^{\hat{g}}P_\mathcal{O},\, ^{\hat{g}}P_{\mathcal{S}},\, ^{\hat{g}}\hat{P}^*_{\mathcal{O}, t},\, t\right) \rightarrow \bigl(\hat{\phi}_{t-1},\, \hat{\rho}_{t-1}\bigr)$, where $t$ is the diffusion timestep.
    \item \textit{Rigid transformation estimation} for rigid objects: We recover an SE(3) alignment between the input object point cloud and the predicted goal configuration using a RANSAC-SVD procedure. We direct the readers to Appendix III-E for implementation details.
\end{enumerate}

\subsection{Global Placement Initialization}
\label{sec:Global Placement Initialization}

%Point cloud diffusion models often focus on single-object point clouds~\cite{luo2021diffusion, Zhou_2021_ICCV, nichol2022point} which are often normalized to a fixed size $[-1, 1]^3$ to match the diffusion prior which noises the point cloud to $\mathcal{N}(0,I)$. However, in the object placement setting, there are two scales to consider: the scale of the object being placed, and the scale of the scene in which the object is being manipulated, which can be arbitrarily large relative to the object itself. Our intuition (based on our experiments) is that these two scales create a conflict: if we normalize the point cloud based on the scale of the object, then the model will have difficulty to diffuse over the entire scene all of the multimodal placement locations.  On the other hand, if we normalize the point cloud based on the scale of the scene, then the model will have difficulty estimating the precise pose of the object in the goal configuration.

Point cloud diffusion models typically focus on generating a single-object in isolation~\cite{luo2021diffusion, Zhou_2021_ICCV, nichol2022point}.  The object is often normalized to a fixed size $[-1, 1]^3$ to match the diffusion prior that noises the point cloud towards $\mathcal{N}(0,I)$. In the object placement setting, however, two scales must be considered: the size of the manipulated object and the size of the surrounding scene, which can be arbitrarily large relative to the object itself. Our intuition, supported by empirical results, suggests that this scale discrepancy introduces a fundamental conflict: normalizing by the object scale hinders the model’s ability to capture multi-modal placement distributions across the full scene, whereas normalizing by the scene scale reduces the model’s capacity to recover the object’s precise goal pose.

%scene-scale diffusion can greatly reduce precision for the placement of relatively small objects. 

To mitigate this issue, we propose a two-stage method: we first predict a local coordinate system (represented by the approximate placement frame $\hat{g}$), and then we predict the object pose in the reference frame of $\hat{g}$. %, which we sample from a distribution of valid placement frames given a scene $\mathcal{S}$. 
Concretely, we consider the centroid of the object in any goal configuration $\mu = \overline{P}^*_{\mathcal{O}}$ to be a valid placement frame, and we aim in the first stage to learn a distribution $f_{\text{global}}(P_{\mathcal{O}}, P_{\mathcal{S}})$ over such placement frames.

To model such a distribution, we learn a feedforward network $f_{\text{global}}$ (see Fig.~\ref{fig:system-local}, \textit{left}) to output a spatially-grounded \emph{Dense Gaussian Mixture Model (GMM)}, in which we predict one Gaussian for each point in the scene $\mathcal{S}$. In particular, given the object and scene point clouds $(P_{\mathcal{O}}, P_{\mathcal{S}})$ as input, we predict for each scene point $p_i \in P_{\mathcal{S}}$ a mixing weight $w_i \in \mathbb{R}$ and a residual vector $r_i \in \mathbb{R}^3$, where  $p_i + r_i$ is the mean of the Gaussian corresponding to point $p_i$. At inference, we simply sample $\hat{g} \sim f_{\text{global}}(P_{\mathcal{O}}, P_{\mathcal{S}})$ from a categorical distribution over the Gaussian means $\{p_i + r_i\}^{N_{\mathcal{S}}}_{i = 1}$ parameterized by the mixing weights $\{ w_i\}^{N_{\mathcal{S}}}_{i = 1}$, similar to standard GMMs. To train $f_{\text{global}}$, we use a negative log-likelihood loss computed using the learned mixing weights:
\begin{align}
\mathcal{L}_{\text{global}}(\mu) 
&= - \log \sum_{i=1}^{N_{\mathcal{S}}} w_i 
    \exp\left(-\tfrac{1}{2\sigma^2} \|p_i + r_i - \mu\|^2\right)
\end{align}
where $\mu = \overline{P}^*_{\mathcal{O}}$ is a ground-truth placement frame. In principle, the variances can also be learned, although we found this detrimental to training stability and unnecessary for approximate placement initialization. 

%\lyuxing{some comparison with Anyplace high-level. Can be dropped is space is limited} This initial placement prediction can be framed in different ways. In a concurrent hierarchical goal prediction framework, AnyPlace~\cite{zhao2025anyplace} utilizes a Vision-Language Model (VLM) for global placement initialization by conditioning on an RGB image and a language prompt to predict a 2D pixel, which is subsequently projected into a 3D coordinate via depth information. In contrast, our geometry-first approach offers several key advantages over such methods. By operating directly on point clouds, our model reasons holistically in 3D space, allowing it to identify valid placements even in regions occluded from a 2D camera viewpoint. This focus on underlying structure, rather than visual appearance, also provides greater robustness to domain shifts in texture and lighting that can degrade the performance of image-based VLMs. Furthermore, while the 2D-to-3D projection of a single pixel is inherently limited and fails to capture placement ambiguity arising from imprecise language prompting, our dense GMM represents a rich, multi-modal probability distribution over all potential goal locations, yielding a more nuanced and geometrically-grounded initialization.

\subsection{Local Configuration Refinement}
\label{sec:Local Placement Refinement}

\noindent\textbf{Disentangled point diffusion.} Our global placement initialization (Sec.~\ref{sec:Global Placement Initialization}) predicts a single point $\hat{g}$ which approximates the centroid of the object in the goal configuration $P^*_{\mathcal{O}}$. This prediction is largely adequate to resolve the placement multi-modality induced by the geometry of a scene $\mathcal{S}$ (e.g., selecting one of multiple mug-racks, or one of multiple pegs on a mug-rack), but cannot capture the precise geometric relationships needed to solve precise manipulation tasks (e.g. the object's exact position and orientation).

In order to estimate the precise object pose in the goal configuration, we need to estimate two things: (i) exactly where the object will be placed, i.e. translation, and (ii) the configuration of the object in the placement pose, i.e. rotation for a rigid object, or shape deformation for a deformable object (as we demonstrate in Appendix I on the project website).

Therefore, we propose \emph{Disentangled Point Diffusion} that disentangles the objective of predicting goal object configuration in point space into diffusing the object frame (i.e. translation) and diffusing the object shape in the goal configuration (i.e. rotation or object deformations).  We express the ground-truth goal configuration as a sum of a mean-centered \emph{shape} $\phi_0$ and a \emph{frame} $\rho_0$:
\begin{align}
\phi_0 \coloneqq {^{\hat{g}}{P^*_{\mathcal{O}}}} - {^{\hat{g}}{\overline{P}^*_{\mathcal{O}}}} \in \mathbb{R}^{N_{\mathcal{O}}\times 3},
\quad
\rho_0 \coloneqq {^{\hat{g}}{\overline{P}^*_{\mathcal{O}}}} \in \mathbb{R}^{3}
\end{align}
where ${^{\hat{g}}{P^*_{\mathcal{O}}}}$ is the ground-truth goal configuration of object $\mathcal{O}$ in frame ${\hat{g}}$, and ${^{\hat{g}}{\overline{P}^*_{\mathcal{O}}}}$ is the centroid of the object in the ground-truth goal configuration in frame ${\hat{g}}$ (i.e. the mean across the $N_{\mathcal{O}}$ object points). These terms are defined so that the ground-truth goal configuration can be computed as ${^{\hat{g}}{P^*_{\mathcal{O}}}}=\phi_0+\rho_0$, where $\rho_0 \in \mathbb{R}^3$ is broadcast to all $N_{\mathcal{O}}$ points during the addition. %Intuitively, the shape $\phi$ captures dense, local transformations (e.g. point displacements from orientation changes or deformations), whereas $\rho$ corrects global translation arising from the approximation error in $\hat{g}$. 
During diffusion noising and denoising, we similarly decompose the goal configuration: 
%we maintain the same composition for running estimates, 
$^{\hat{g}}\hat{P}^*_{\mathcal{O},t}=\hat{\phi}_t+\hat{\rho}_t$, again broadcasting $\hat{\rho}_t$ across object points.

We model two decoupled forward corruption processes with a shared noise schedule:
\begin{align}
\phi_t &= \sqrt{\bar{\alpha}_t}\,\phi_0 \;+\; \sqrt{1-\bar{\alpha}_t}\,\epsilon_\phi, 
\quad \epsilon_\phi \sim \mathcal{N}(0,I) \label{eq:shape-diffusion}\\
\rho_t &= \sqrt{\bar{\alpha}_t}\,\rho_0 \;+\; \sqrt{1-\bar{\alpha}_t}\,\epsilon_\rho,
\quad \epsilon_\rho \sim \mathcal{N}(0,I) \label{eq:frame-diffusion}
\end{align}
Both Eqns~\ref{eq:shape-diffusion} and~\ref{eq:frame-diffusion} apply per-point isotropic Gaussian noising following prior work~\cite{luo2021diffusion, Zhou_2021_ICCV} under the same schedule $\{\bar{\alpha}_t\}_t$. %In principle, component-specific (scale-aware) schedules could be more optimal for $\phi$ and $\rho$ individually; 
% For general applicability across objects and scenes, we first normalize the object and scene point clouds by object scale and then adopt the standard Gaussian corruption for both components, assuming (in normalized coordinates) that $\phi$ and $\rho$ are calibrated to the corruption model—i.e., approximately unit-scale, isotropic signals whose residuals are well matched to $\mathcal{N}(0,I)$. This ensures that the schedule $\{\bar{\alpha}_t\}_t$ induces a well-conditioned signal-to-noise ratio for both components across timesteps, so corruption proceeds at a comparable rate (neither too fast nor too slow) for shape and frame alike.
The reverse process is estimated by a dual-head denoiser operating in the local frame $\hat{g}$, i.e. $f_{\text{local}}\!\left(^{\hat{g}}P_\mathcal{O},\, ^{\hat{g}}P_{\mathcal{S}},\, ^{\hat{g}}\hat{P}^*_{\mathcal{O}, t},\, t\right) \rightarrow \bigl(\hat{\phi}_{t-1},\, \hat{\rho}_{t-1}\bigr)$, where the inputs are the object and scene point clouds and the current estimate at denoising step $t$ of the object in the goal configuration in frame $\hat{g}$: $^{\hat{g}}\hat{P}^*_{\mathcal{O}, t}=\hat{\phi}_t+\hat{\rho}_t$. The network directly predicts the next-step disentangled estimates $(\hat{\phi}_{t-1}, \hat{\rho}_{t-1})$, which are composed to yield the updated goal configuration $^{\hat{g}}\hat{P}^*_{\mathcal{O}, t-1} \;=\; \hat{\phi}_{t-1} + \hat{\rho}_{t-1}$. Next we describe the inputs to the network that represents $f_{\text{local}}$, shown in Fig.~\ref{fig:system-local} (\textit{right}).

\noindent\textbf{Reconstruction embedding.} Rather than encoding the denoised object and the scene separately (similar to~\cite{cai2024tax3d, simeonov2023shelving, zhao2025anyplace}), we input to the diffusion model the \textit{reconstructed} placement point cloud, consisting of the scene $^{\hat{g}}P_{\mathcal{S}}$ combined with the current denoised object in the predicted goal configuration $^{\hat{g}}\hat{P}^*_{\mathcal{O}, t}$. This combined point cloud $(^{\hat{g}}P_{\mathcal{S}}, ^{\hat{g}}\hat{P}^*_{\mathcal{O}, t})$ is processed by a single point cloud encoder to obtain per-point \textit{reconstruction embeddings} $\{f_i\}^{N_\mathcal{O} + N_\mathcal{S}}_{i=1}$, allowing the network to more precisely model the geometric relationship between the object and the scene. We also separately compute per-point \textit{object embeddings} $\{o_j\}^{N_\mathcal{O}}_{j=1}$ by encoding the initial, mean-centered object point cloud ${^{\mathcal{O}}}P_{\mathcal{O}} = P_{\mathcal{O}} - \bar{P}_{\mathcal{O}}$.  These object embedding tokens are used to predict the goal shape $\phi$, which is also mean-centered.

\noindent\textbf{Deformation embedding.} To reason more granularly about how object $\mathcal{O}$ must transform, $f_{\text{local}}$ contains an additional deformation encoder that takes as input the per-point displacements $\hat{\phi}_t - {^{\mathcal{O}}}P_{\mathcal{O}}$ between the current denoised and initial object shapes, and computes a per-point \textit{deformation embedding} $\{d_k\}^{N_\mathcal{O}}_{k=1}$. As $\hat{\phi}_t$ and ${^{\mathcal{O}}}P_{\mathcal{O}}$ are both mean-centered, their difference explicitly models local transformations due to rotations or deformations.

\noindent\textbf{Rotation noise.} %Following~\cite{luo2021diffusion, Zhou_2021_ICCV}, we add per-point Gaussian noise to the shape $\phi_0$ during the forward process, i.e. $\phi_t = \sqrt{\bar{\alpha}}_t\phi_0 + \sqrt{1 - \bar{\alpha}_t}\epsilon$, where $\epsilon \sim \mathcal{N}(0, I)$. 
To enable $f_{\text{local}}$ to more precisely denoise pose transformations, we sample an additional rotation noise term $\epsilon_{\text{rot}} = R\phi_0 - \phi_0, \text{ where } R$ is sampled from a distribution over $\mathrm{SO}(3)$. Consequently, our forward diffusion process for the shape $\phi_0$ during training is: $\phi_t = \sqrt{\bar{\alpha}}_t\phi_0 + \sqrt{1 - \bar{\alpha}_t}(\epsilon_\phi + \epsilon_{\text{rot}})$. The reverse process for the shape $\phi$ remains unchanged. Details for the rotation noise implementation can be found in Appendix III-B on the project website. 

Given the sampled global placement reference frame $\hat{g}$ and the denoised local placement configuration ${^{\hat{g}}\hat{P}^*_\mathcal{O}}$, we can then compute the global predicted goal placement point cloud as $\hat{P}^*_\mathcal{O} = {^{\hat{g}}\hat{P}^*_\mathcal{O}} + \hat{g}$.  The robot can then move the object to this pose, using either a learned goal-conditioned policy (Appendix I-B) or by estimating an SE(3) transformation (Appendix III-E) followed by motion planning.

\begin{table*}[!t]
    \centering
    \scriptsize
    \caption{Ablations and Task Success Rates on RPDiff tasks.}%RPDiff %results are reported from its original paper; that method uses heuristic local cropping and a uses separate success classifier, whereas TAX-DPD is evaluated one-shot.}
    \label{tab:results_table_sim}
    \begin{tabular}{l|ccccc|c}
    \toprule
         & \texttt{Mug/EasyRack} & \texttt{Mug/MedRack} & \texttt{Mug/Multi-MedRack} & \texttt{Book/Shelf} & \texttt{Can/Cabinet} & \texttt{Average}\\
    \midrule
        TAX3D~\cite{cai2024tax3d} & 0.84 & 0.46 & 0.32 & 0.38 & 0.42 & 0.48\\
        RPDiff~\cite{simeonov2023shelving} (w/o classifier-based reranking) & - & - & - & - & - & 0.83 \\
        RPDiff~\cite{simeonov2023shelving} & 0.92 & 0.83 & 0.86 & 0.94 & 0.85 & 0.88 \\
    \midrule
        \textit{TAX-DPD (Ours)} w/o disentangled point diffusion & 0.97 & 0.74 & 0.61 & 0.53 & 0.77 & 0.72\\
        \textit{TAX-DPD (Ours)} w/ MLP encodings & 0.99 & 0.84 & 0.81 & 0.61 & 0.64 & 0.78\\
        \textit{TAX-DPD (Ours)} w/o GMM & \textbf{1.00} & 0.87 & 0.74 & 0.75 & 0.79 & 0.83 \\
        \textit{TAX-DPD (Ours)} w/o recon. embedding & 0.98 & 0.91 & 0.80 & 0.78 & 0.83 & 0.86 \\
        \textit{TAX-DPD (Ours)} w/o rot. noise & 0.94 & 0.85 & 0.73 & 0.96 & 0.91 & 0.88\\
        \textit{TAX-DPD (Ours)} w/o deform. embedding & 0.98 & 0.94 & 0.88 & 0.95 & 0.80 & 0.91 \\
        \textit{TAX-DPD (Ours)} w/ SE(3) diffusion & 0.97 & 0.92 & 0.89 & 0.96 & 0.91 & 0.93 \\
    \midrule
        \textit{TAX-DPD (Ours)} & \textbf{1.00} & \textbf{0.97} & \textbf{0.95} & \textbf{0.99} & \textbf{0.95} & \textbf{0.97}\\
    \bottomrule
    \end{tabular}
\end{table*}

\begin{comment}
\subsection{Estimating Relative Transformation}
\label{sec:estimating relative transformation}
Given the sampled global placement reference frame $\hat{g}$ and the denoised local placement configuration ${^{\hat{g}}\hat{P}^*_\mathcal{O}}$, we can compute the global predicted goal placement point cloud as $\hat{P}^*_\mathcal{O} = {^{\hat{g}}\hat{P}^*_\mathcal{O}} + \hat{g}$. %Since our model design embeds per-point correspondence as an inductive bias, 
In our model design, each input point $p_i \in P_\mathcal{O}$ directly corresponds to a predicted goal point $\hat{p}^*_i \in \hat{P}^*_\mathcal{O}$.  

For some downstream applications, we may need to predict the final SE(3) pose of the target object.
However, since our diffusion process denoises points independently, the predicted goal set may contain local inconsistencies and outliers that violate rigid-body constraints. To address this, we employ a RANSAC-SVD approach as a  post-processing step that projects the noisy correspondences onto a single rigid-body transform. Specifically, RANSAC iteratively:
\begin{enumerate}
    \item samples a set of three correspondences $\{(p_i, \hat{p}^*_i)\}$, 
    \item estimates a candidate transformation $T$ from these correspondences, 
    \item evaluates inlier support by counting correspondences that satisfy $\|T p_j - \hat{p}^*_j\|_2 < \tau$, where $\tau$ is a distance threshold.
\end{enumerate}
After $N$ iterations, we select the transform with the largest inlier set, then re-estimate the final SE(3) transform using SVD over all inliers.
\end{comment}

\subsection{Additional Architecture \& Training Details.} %Mention PointNet++, and modified point DiT with cross-attention here. Also need to make some mention of how we train with $\hat{g}$ without querying the GMM model.
We compute reconstruction, object, and deformation embeddings each with a PointNet++~\cite{qi2017pointnet++} point cloud encoder. These embeddings are aggregated into object and scene tokens, and are passed as input along with a learnable frame token into a modified Diffusion Transformer (DiT)~\cite{peebles2023scalable}, where the object tokens and frame token cross-attend to the scene tokens. The frame token is then decoded by the frame prediction head into $\hat{\rho}_t$, and the object tokens are decoded by the shape prediction head into $\hat{\phi}_t$.  The current denoised configuration $^{\hat{g}}\hat{P}^*_{\mathcal{O}, t} = \hat{\rho}_t + \hat{\phi}_t$ is passed back as input into the model for the next diffusion timestep.

%For efficiency, we do not sample $\hat{g}$ from $f_{\text{global}}$ for training $f_\text{local}$. Given some target goal configuration $P^*_{\mathcal{O}}$, we instead sample $\hat{g}$ by adding noise to the ground-truth placement frame $\overline{P}^*_{\mathcal{O}}$, i.e. $\hat{g}_{\text{train}} \sim \mathcal{N}(\overline{P}^*_{\mathcal{O}}, \Sigma)$. For simplicity, we use $\Sigma = I$, though this parameter can be tuned to match the scale of the errors in $f_{\text{global}}$. Additional architecture and training details can be found in Appendix II and III on project website.

For efficiency and stable training, we do not sample $\hat{g}$ from $f_{\text{global}}$ when training $f_\text{local}$. 
Querying $f_{\text{global}}$ for each training sample significantly increases training cost and may introduce mode mismatch, as predictions from $f_{\text{global}}$ may fall into a different mode than the ground-truth placement $P^*_{\mathcal{O}}$, forcing $f_{\text{local}}$ to handle large translational errors rather than perform local shape refinements. Instead, we sample $\hat{g}$ by adding noise to the ground-truth placement frame $\overline{P}^*_{\mathcal{O}}$, i.e. $\hat{g}_{\text{train}} \sim \mathcal{N}(\overline{P}^*_{\mathcal{O}}, \Sigma)$. For simplicity, we use $\Sigma = I$, though this parameter can be tuned to match the scale of the errors in $f_{\text{global}}$. In practice, the global stage only needs to place $\hat{g}$ inside the basin where local refinement is effective, so this training scheme remains well aligned with the test-time objective while avoiding unnecessary global ambiguity. Additional architecture and training details can be found in Appendix II and III on the project website.

\section{Experiments}
We include experiments in both simulation on standard object placement benchmarks (Sec.~\ref{sec:simulation_experiments}) as well as real-world insertion for manufacturing-related tasks using the NIST-board (Sec.~\ref{sec:realworld_experiments}). TAX-DPD can theoretically be applied to placement for non-rigid objects, since it makes no assumptions about SE(3) rigidity.  See Appendix I for simulation results of placing non-rigid cloths on hangers.  However, non-rigid placement in the real world is challenging due to the difficulty of 3D tracking of non-rigid objects; we leave handling of these issues for future work. We direct the reader to our project website for supplementary materials and video demonstrations.

\subsection{Simulation Experiments}
\label{sec:simulation_experiments}
\subsubsection{Experimental Setup} Our simulation experiments are conducted on the full suite of RPDiff \cite{simeonov2023shelving} placement tasks, which are implemented in the PyBullet \cite{bullet} simulation engine and designed to evaluate precise relational object rearrangement in complex, multi-modal environments. The suite of tasks includes \texttt{Mug/EasyRack}, \texttt{Mug/MedRack}, \texttt{Mug/Multi-MedRack}, \texttt{Book/Shelf}, and \texttt{Can/Cabinet}, collectively spanning different degrees of placement precision, multi-modality, and geometric variations. Each task has the following objective: (1) hanging a mug on one rack with one peg, (2) hanging a mug on one rack with two pegs, (3) hanging a mug on multiple racks with two pegs, (4) inserting a book into a partially filled bookshelf, (5) stacking a can on top of a stack of cans or onto an open shelf. For detailed descriptions and visualizations of the RPDiff tasks, please refer to the RPDiff paper \cite{simeonov2023shelving}.

\subsubsection{Evaluation and Metrics} We adopt the insertion controller introduced in RPDiff to execute the predicted placements produced by TAX-DPD and the baselines. For each placement task, we evaluate success rates over 100 trials, where in each trial the scene configuration is generated by spawning randomly sampled meshes with randomized poses for both placement and scene objects from a held-out test suite (unseen during training). Success is determined by evaluating the final simulator state after the placement. %, assessing whether the object is placed correctly relative to the scene after the insertion controller executes the predicted placement pose. 
%For the \texttt{Book/Bookshelf} task, we further quantify multi-modal capture and placement precision by comparing the set of predicted placements against a ground-truth set of feasible solutions and reporting corresponding precision and recall.

\subsubsection{Baselines}
We compare TAX-DPD to \textbf{RPDiff}~\cite{simeonov2023shelving} and \textbf{TAX3D}~\cite{cai2024tax3d}, two recent diffusion-based approaches that operate on distinct domains. RPDiff models object rearrangement as an iterative de-noising process directly in the space of rigid SE(3) transformations, which guides a perturbed transformation toward a valid placement using object-scene point clouds. To enhance generalization and precision, RPDiff crops a local point cloud context around the object based on heuristics and employs a separately trained success classifier to select the highest-scoring prediction for evaluation. We report RPDiff's success rates presented in its original paper. In contrast, TAX3D operates as a diffusion model in 3D point space, predicting dense displacements for the object conditioned on the scene. Although originally designed for deformable objects, we train it on the same task suite and apply an identical rigid transformation estimation procedure (Appendix III-E) for a fair comparison.

%\noindent\textbf{RPDiff}~\cite{simeonov2023shelving} models object rearrangement as an iterative pose de-noising process on object–scene point clouds, allowing it to iteratively guide a perturbed SE(3) transformation toward a multi-modal distribution of valid placements. Notably, to enhance generalization and precision, RPDiff crops a local point cloud context around the object based on heuristics and employs a separately trained success classifier to select the highest-scoring prediction for evaluation. We report RPDiff's success rates presented in its original paper.

%\noindent\textbf{TAX3D}~\cite{cai2024tax3d} extends relative placement to deformable objects by predicting dense, per-point displacements through a point cloud diffusion process conditioned on both object and scene point clouds. We train it on the same RPDiff task suite and evaluate it using the same rigid transformation estimation procedure (Sec.~\ref{sec:estimating relative transformation}) as our own method for fair comparison.

\begin{table}[t]
    \centering
    \scriptsize
    \caption{Additional Ablations Comparing Point Diffusion to SE(3) Diffusion on variations of the Mug Hanging task.}
    \label{tab:additional_ablate}
    \begin{tabular}{l|cc}
    \toprule
         & \texttt{OneMug} & \texttt{ManyMugs} \\
    \midrule
        \emph{TAX-DPD (Ours)} w/ SE(3) diffusion & 0.97 & 0.89 \\
        \emph{TAX-DPD (Ours)} (Point diffusion) & \textbf{0.98} & \textbf{0.95} \\
    \bottomrule
    \end{tabular}
\end{table}

\subsubsection{Comparison between SE(3) diffusion and Point Cloud Diffusion}
Table~\ref{tab:results_table_sim} reports success rates across the RPDiff task suite for baselines (top rows), our ablation variants (middle rows), and TAX-DPD (bottom row). %We focus our analysis on the baselines and our full method, which establishes a new state of the art by demonstrating a level of placement \emph{precision} unmet by prior work.
TAX-DPD demonstrates superior performance over the RPDiff baseline, which diffuses in SE(3) space. We achieve an average success rate of 97\%, establishing a significant 9\% margin over RPDiff's 88\%. Specifically, in tasks characterized by significant geometric variations across unseen objects (e.g. \texttt{Mug/EasyRack}, \texttt{Mug/MedRack}, and \texttt{Mug/Multi-MedRack}), TAX-DPD consistently generates more feasible placements under these shape variations; in cluttered scenes requiring high precision (e.g. \texttt{Book/Shelf} and \texttt{Can/Cabinet}), our approach demonstrates a superior capacity for fine-grained local geometric reasoning, without requiring heuristic local cropping adopted in RPDiff. %that cannot scale to general applications. 
Furthermore, RPDiff trains a separate learned classifier to score sampled poses and takes the highest-ranking prediction for evaluation, while TAX-DPD directly evaluates sampled configurations in one-shot, without heuristic local cropping or classifier-based reranking. Compared to RPDiff without a classifier, which is a more direct comparison, our approach has an even larger performance gain of 14\%.

To isolate the benefits of diffusing in point space from other architectural choices, 
we create a controlled ablation by adapting our own method to perform diffusion directly 
on the SE(3) manifold (See Appendix III-C for implementation details). As shown in Table~\ref{tab:results_table_sim}, this variant, dubbed ``\textit{TAX-DPD (Ours)} w/ SE(3) diffusion'', underperforms our full method by an average of 4\%, with this deterioration being most significant on mug-hanging tasks involving substantial object geometric variations. This result confirms that operating in point space is a key contributor to our model's success.

To further analyze \emph{why} point cloud diffusion is  preferable, % and validate our insight 
%that it better handles geometric variations, 
we conduct a targeted analysis on two versions of the \texttt{Mug/Multi-MedRack} task. The first, which we term \texttt{OneMug}, uses a fixed mug geometry and a fixed set of racks to isolate the placement challenge from shape variation. The second, \texttt{ManyMugs}, is the original task featuring diverse, unseen geometries for both mugs and racks. As shown in Table~\ref{tab:additional_ablate}, the two methods perform comparably on \texttt{OneMug} (98\% vs. 97\%) when the object geometry has no variations. However, when faced with the diverse, unseen shapes of the \texttt{ManyMugs} task, TAX-DPD's performance remains robust with only a small drop ($98\%\rightarrow$95\%), whereas the SE(3) variant suffers a more significant deterioration ($97\%\rightarrow$89\%). This aligns with our insight that point cloud diffusion, in contrast to SE(3) diffusion, supports reasoning about low-level object geometry, thereby facilitating more effective generalization to geometric variations in object manipulation.

\begin{table*}[t]
    \centering
    \scriptsize
    \caption{Task Success Rates and Precision Metrics on Real-World Insertion Tasks. For the multimodal Waterproof task, translation and rotation errors are omitted because there is no single canonical target pose.}
    \label{tab:results_table_real}
    \begin{tabular}{lcccccc|cc}
    \toprule
         & \multicolumn{6}{c}{Unimodal} & \multicolumn{2}{c}{Multimodal} \\
         \cmidrule(lr){2-7} \cmidrule(lr){8-9}
         & \multicolumn{2}{c}{Waterproof} 
         & \multicolumn{2}{c}{DSUB-25} 
         & \multicolumn{2}{c}{SSD} 
         & \multicolumn{2}{c}{Waterproof} \\
         \cmidrule(lr){2-3} \cmidrule(lr){4-5} \cmidrule(lr){6-7} \cmidrule(lr){8-9}
         & TAX-Pose & \textit{TAX-DPD (Ours)} 
         & TAX-Pose & \textit{TAX-DPD (Ours)} 
         & TAX-Pose & \textit{TAX-DPD (Ours)}
         & \textit{TAX-DPD (Ours)} \\
    \midrule
         Success Rate 
         & 80\% (16/20) & \textbf{100\% (20/20)} 
         & \textbf{80\% (16/20)} & \textbf{80\% (16/20)} 
         & 0\% (0/20) & \textbf{85\% (17/20)} 
         & \textbf{90\% (18/20)} \\
         Trans. Err. (mm) 
         & 1.04 & \textbf{0.72} 
         & \textbf{0.93} & 1.00 
         & 16.18 & \textbf{2.75} 
         & - \\
         Rot. Err. ($^\circ$) 
         & 1.64 & \textbf{1.18} 
         & 3.16 & \textbf{1.36} 
         & 13.81 & \textbf{2.77} 
         & - \\
    \bottomrule
    \end{tabular}
\end{table*}

\subsubsection{Benefits of Global Placement Initialization}

\begin{figure}[t]
   \centering
   \includegraphics[width=\linewidth]{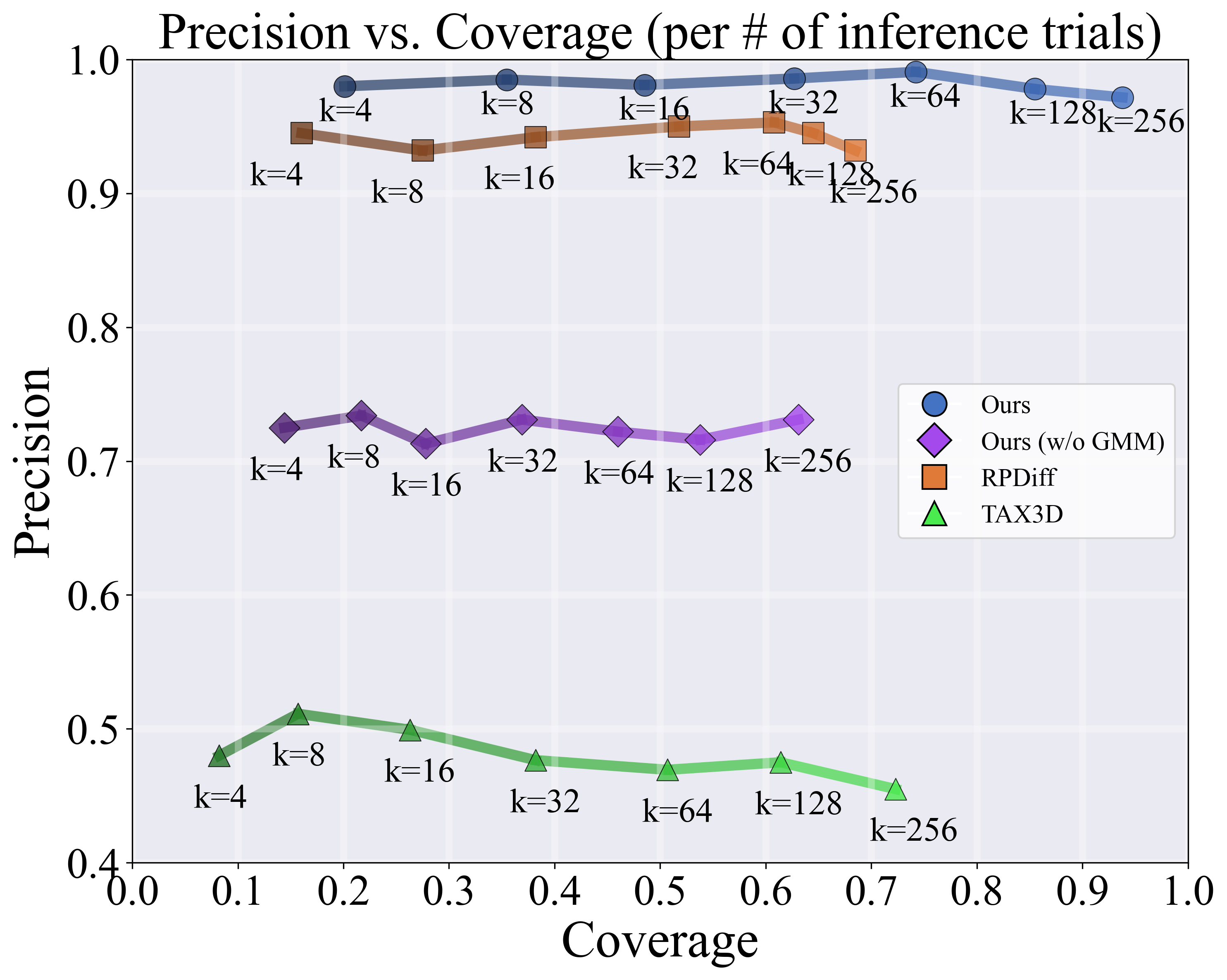}
   \caption{\textbf{Coverage vs. Precision.} We further evaluate TAX-DPD and the baselines on coverage and precision with increasing numbers of inference samples on the RPDiff task \texttt{Book/Shelf}.}
   \label{fig:coverage_precision}
\end{figure}

To understand the importance of global placement initialization, we also perform an ablation in which we remove the dense GMM and instead initialize the diffusion process at the centroid of the scene point cloud (``\textit{TAX-DPD (Ours)} w/o GMM").  Table~\ref{tab:results_table_sim} shows that this leads to a drop in performance of 14\%.  %To further analyze this, 
We further compute the coverage and precision of the \texttt{Book/Shelf} task for different values of $K$, which is defined as the number of predictions sampled %for each method 
(see Figure~\ref{fig:coverage_precision}). Here, coverage measures the fraction of feasible ground-truths that are within a threshold distance of at least one of the $K$ samples, while precision measures the fraction of the $K$ sampled predictions that are within a threshold distance of one of the ground-truths. TAX-DPD achieves state-of-the-art \emph{coverage} while maintaining high precision.  In contrast, the ablation of our method without the GMM has a significant drop in both precision and coverage, highlighting the importance of the global placement initialization. %(which we achieve using the dense GMM). %to obtain placement coverage over the full scene.  

\subsubsection{Benefits of Disentangled Point Diffusion}
To understand the importance of disentangled point diffusion, we perform an ablation in which we directly predict the goal configuration with point cloud diffusion (``\textit{TAX-DPD (Ours)} w/o disentangled point diffusion''). As shown in Table~\ref{tab:results_table_sim}, removing disentangled point diffusion causes the most performance degradation of 25\%, confirming that forcing the model to predict dense geometry and translation as a combined output is ineffective for point cloud diffusion approaches.

\subsubsection{Ablation Study}
We further ablate the remaining design choices of TAX-DPD in Table~\ref{tab:results_table_sim}. Variants that replace the encoder with simple MLPs or remove reconstruction or deformation embeddings all degrade performance, indicating the need for expressive feature representations that capture fine-grained geometric structure for precise dense prediction. In addition, rotation noise perturbations are essential for learning diverse poses, and their removal lowers the average success rate to 88\%, most notably in mug-hanging tasks.

%\subsubsection{Ablation Study}
%We further ablate the remaining design choices of our method in Table~\ref{tab:results_table_sim}. The comparison indicates that each of our design choices contributes meaningfully to our overall performance. \textit{Removing the disentangled shape-frame diffusion causes the most significant performance degradation of 25\%, confirming that forcing the model to simultaneously learn dense geometry and translations is ineffective.} The ablations further show the importance of using a point-cloud encoder~\cite{qi2017pointnet++} instead of a per-point MLP encoder, using a reconstruction embedding, a deformation embedding, and adding additional rotation noise during training.

%Additionally, our dedicated point cloud encoding scheme is crucial. The poor performance of the variant using simple MLPs (w/ MLP encodings) underscoring that dense, per-point prediction demands a sophisticated encoder capable of capturing fine-grained geometric structure. Removing either the reconstruction or deformation embeddings individually degrades performance, highlighting their importance in modeling object-scene interactions and transformations across the iterative denoising process. Finally, the inclusion of rotation noise perturbations proves vital for learning complex poses, and its removal drops the average success rate to 88\%. This is especially evident in the mug hanging tasks, which demand the model to learn a diverse distribution of valid placement poses.

\begin{figure}[b]
  \centering
  \subfloat[\shortstack{Unimodal\\Waterproof}]{
      \includegraphics[width=0.20\linewidth]{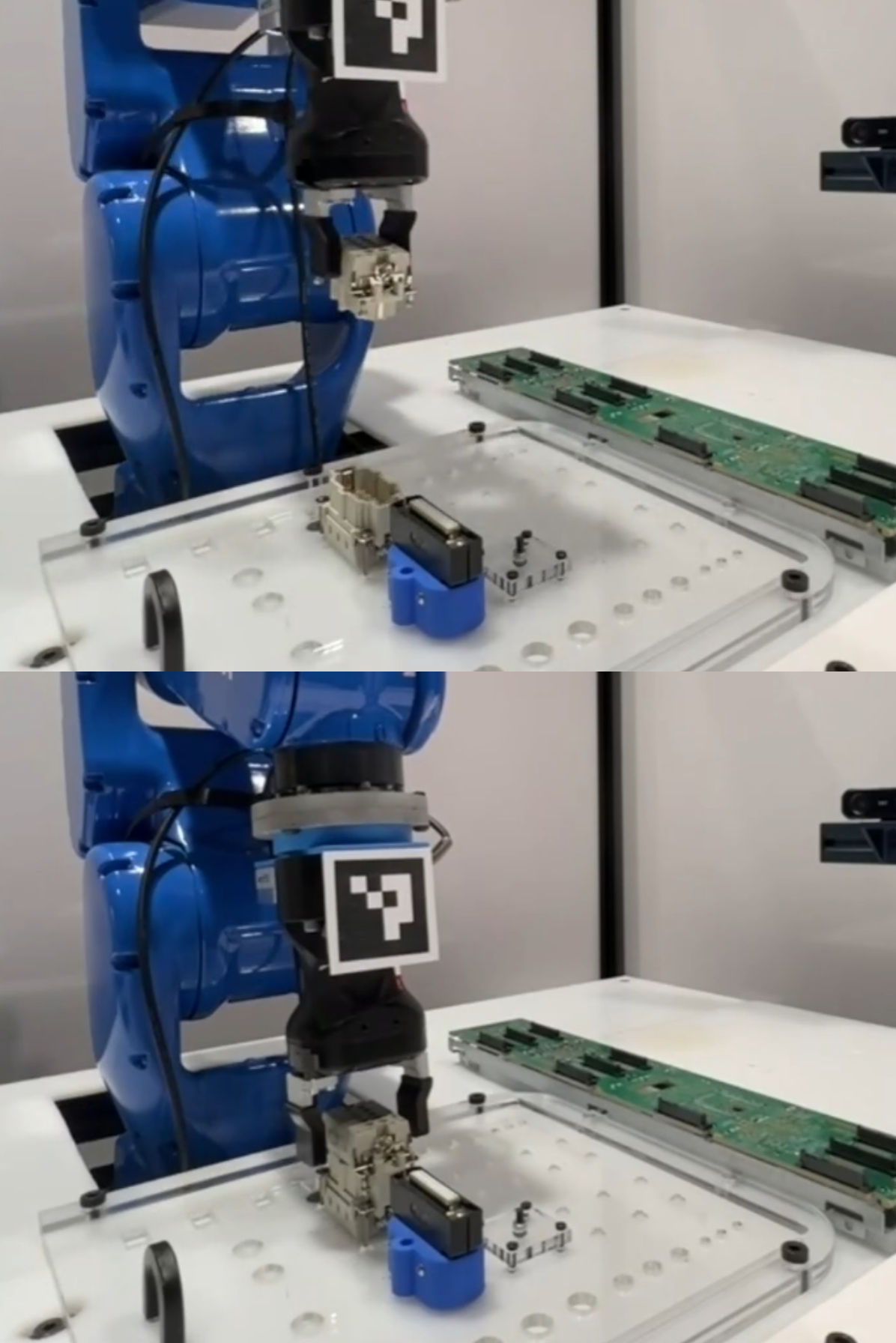}}
  \hfill
  \subfloat[\shortstack{Unimodal\\DSUB-25}]{
      \includegraphics[width=0.20\linewidth]{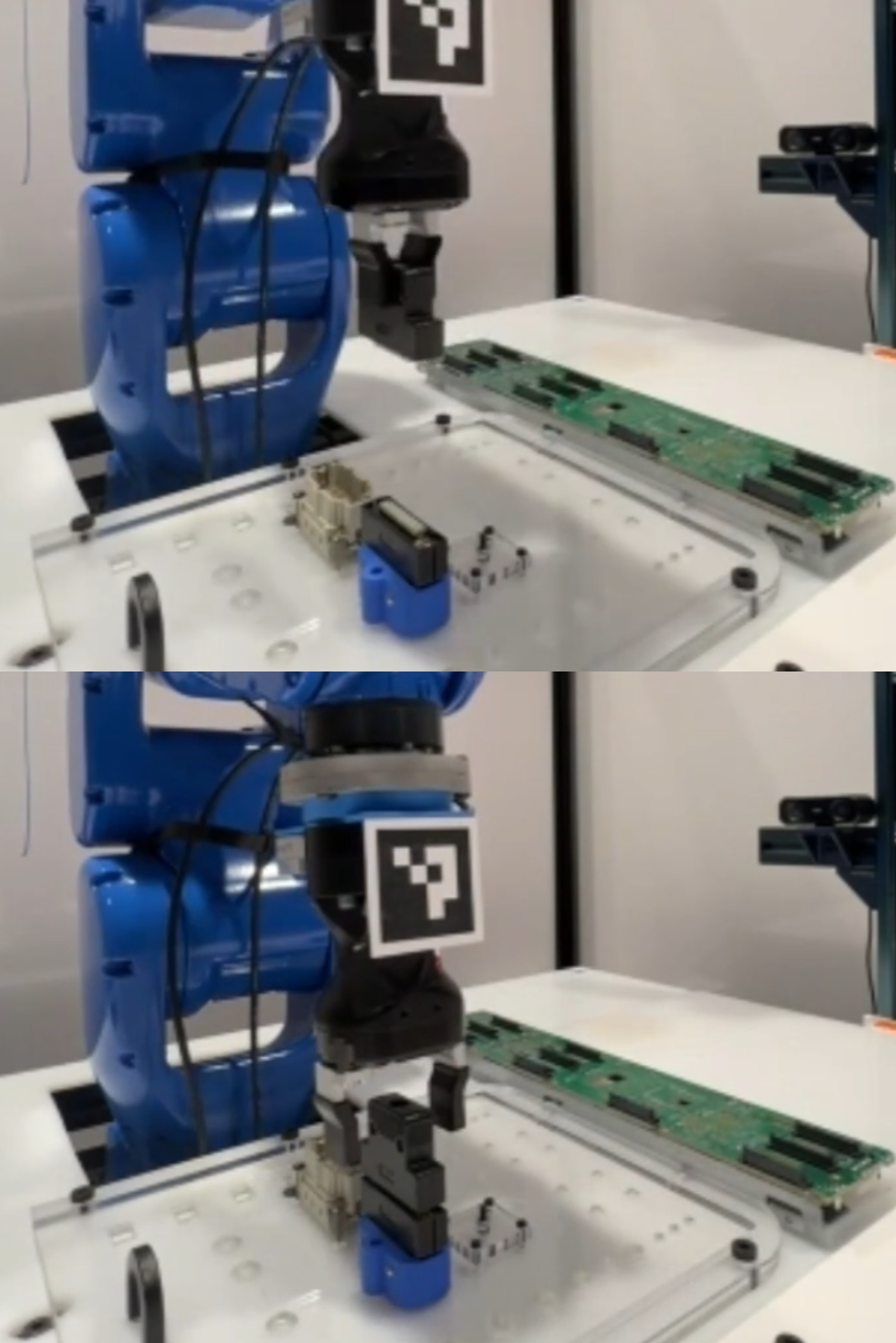}}
  \hfill
  \subfloat[\shortstack{Unimodal\\SSD}]{
      \includegraphics[width=0.20\linewidth]{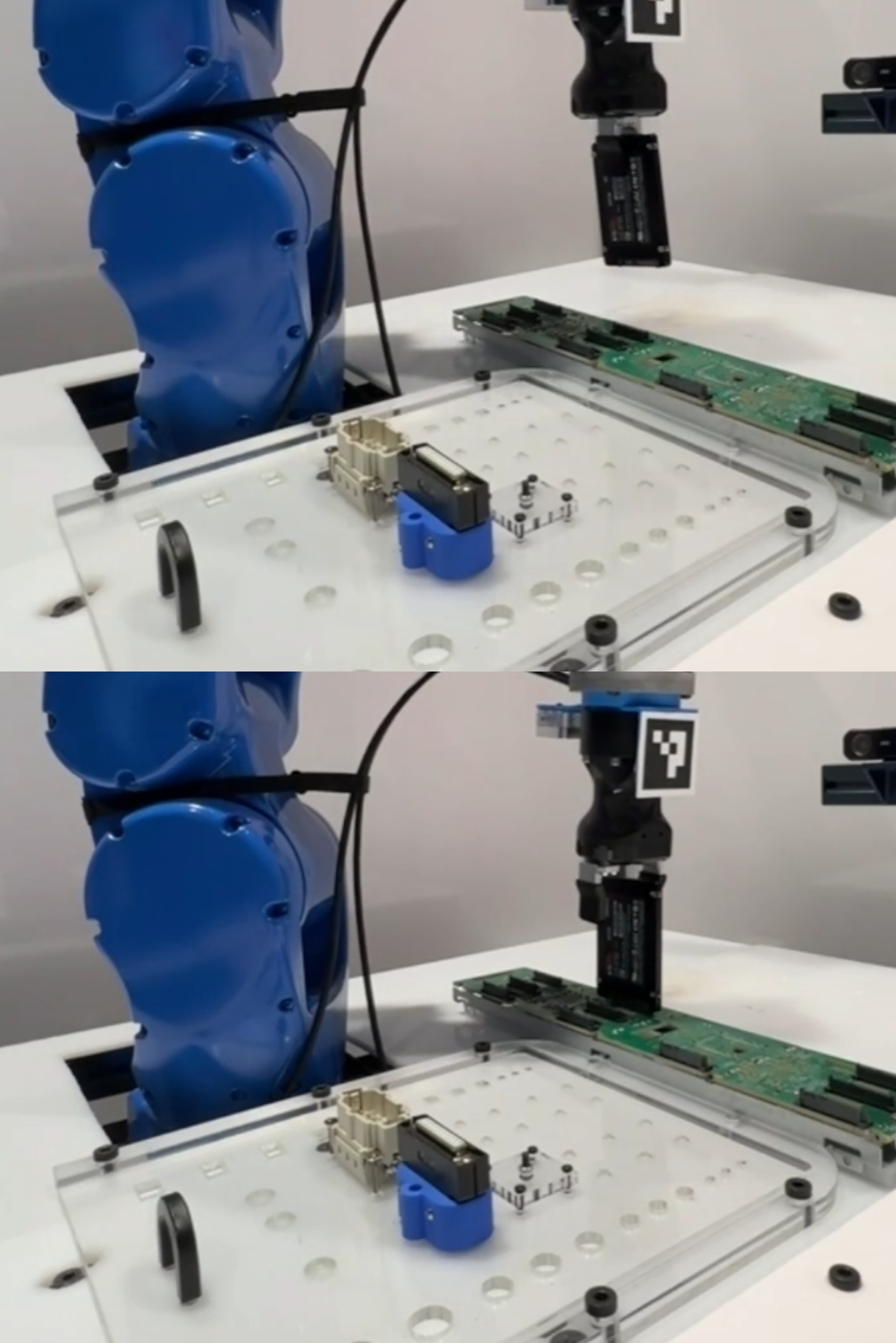}}
  \hfill
  \subfloat[\shortstack{Multimodal\\Waterproof}]{
      \includegraphics[width=0.20\linewidth]{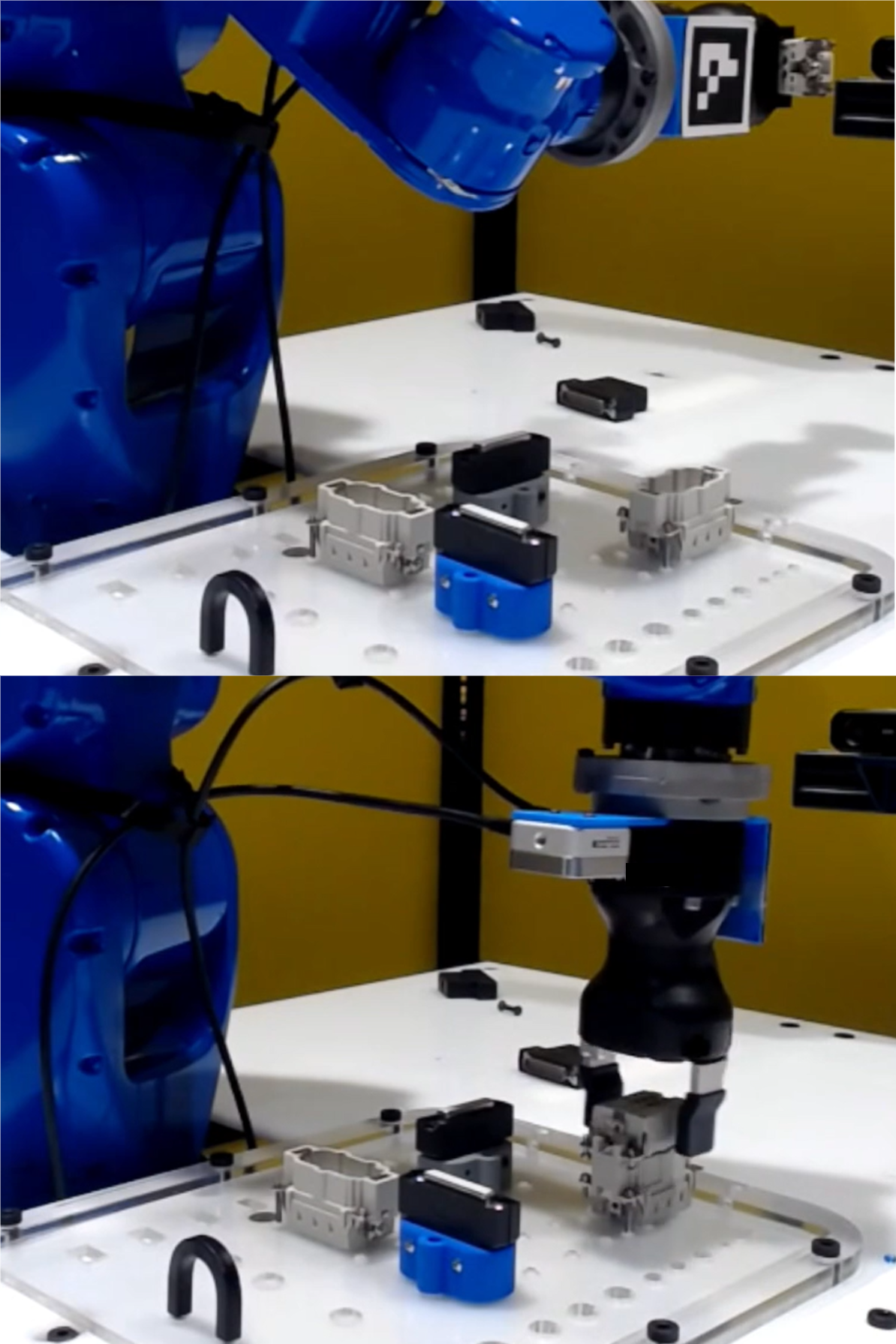}}
  \caption{\textbf{Insertion task rollouts.} Selected TAX-DPD rollouts. Top: pre-insertion. Bottom: post-insertion.}
  \label{fig:insertion_viz}
\end{figure}

\subsection{Real-World Experiments}
\label{sec:realworld_experiments}
\subsubsection{Experimental Setup}
To examine whether TAX-DPD can reliably complete real-world placement tasks, we evaluate it on high-precision industrial tasks with three distinct connectors: the ``Waterproof'', ``DSUB-25'', and ``SSD'' connector from the NIST Assembly Task Board 1 \cite{nist_assembly_2018} (see Figure~\ref{fig:pull} and Figure~\ref{fig:insertion_viz}). Our setup uses a 6-DOF robot arm with a dual-camera system (wrist and side) to capture point clouds of the plug and the sockets.
%We designed unimodal and multi-modal placement tasks using these connectors, illustrated in Figure~\ref{fig:insertion_viz}. For the unimodal tasks, the robot inserts each of the three distinct NIST connectors into its single corresponding socket,  with 
The primary challenge arises from significant, random perturbations to the connector's initial configuration via varying the object's initial pose in the gripper. 
%gripper's starting pose when picking up the connector 
%The multi-modal task focuses on the ``Waterproof'' connector but increases scene complexity by introducing a second valid placement mode and geometrically distinct sockets as distractors. 
For each task, we collect a dataset of 20 demonstrations of successful connection placements via teleoperations. More details about the real-world experiments can be found in Appendix IV.

\subsubsection{Insertion Success Rate Results}
%As seen in Table~\ref{tab:results_table_real}, our method achieves very high levels of success across all four of these difficult insertion tasks, with success rates ranging from 80-100\%.  This task is especially difficult because it requires around millimeter-level precision and around 1-2 degrees of rotational accuracy for successful insertion.

The results presented in Table~\ref{tab:results_table_real} underscore the effectiveness of our approach. We achieved 80-100\% success across four high-precision insertion tasks that necessitate positional accuracy at the millimeter scale and rotational error within a 1-2 degree tolerance, highlighting the difficulty of accomplishing the insertion tasks in an industrial setting. TAX-DPD also matches or dramatically outperforms TAX-Pose~\cite{pan2023tax}, a strong baseline tailored for unimodal relative placement tasks through learned soft correspondences. Notably, TAX-Pose fails all 20 SSD trials, producing extremely high errors. Because the SSD object is relatively tall, small rotation errors in the predicted bottom-face point cloud lead to large errors in the resulting gripper pose. TAX-DPD, on the other hand, is still able to achieve high success rates. Furthermore, in the multi-modal version of the Waterproof connector insertion task, our model still achieves a reliable success rate of 90\%. 

%As seen in Table~\ref{tab:results_table_real}, across all three unimodal insertion tasks, our method matches or dramatically outperforms TAX-Pose. For the Waterproof and DSUB-25 unimodal insertion tasks, both our model and TAX-Pose achieve sub-millimeter translational accuracy and low rotation error, underscoring the demanding precision requirement for successful insertions. Notably, TAX-Pose incurs extremely high errors in the SSD task, failing on all 20 trials. Since the SSD object is relatively tall, small rotation errors in the point cloud prediction of the bottom face of the object can compound into large errors in the predicted gripper pose. Our method, on the other hand, is still able to achieve high success rates. Furthermore, in the challenging multi-modal Waterproof connector insertion task, our model achieves a reliable success rate of 90\%, whereas TAX-Pose, being restricted to unimodal predictions, is unable to handle the task. Importantly, our approach attains this level of precision as a generative model of full multi-modal placement distributions, outperforming TAX-Pose which directly regresses to a single deterministic output. Achieving equal or superior accuracy in a generative regime highlights the robustness and precision of our predictions.

% \section{Extension to Non-Rigid Placement Tasks}

%\lyuxing{mention it is hard to require tracking (and cite tracking papers), and say leave it for future works}

\section{Conclusion}
We present a hierarchical goal prediction framework that pairs a scene-conditioned Dense GMM for global placement with a local point-cloud diffuser that jointly denoises the object geometry and frame in local coordinates. This design resolves scene-level multi-modality while preserving placement precision and generalizes across object geometries. In simulation (RPDiff), we attain state-of-the-art success rates, significantly outperforming prior work.  Our analysis shows that point cloud diffusion significantly outperforms SE(3) diffusion especially when the task requires generalizing over variations in object geometry. %; ablations show the GMM drives coverage and the disentangled diffuser drives fine-grained accuracy, evidencing the robustness of learning in point-space over SE(3) space. 
On a millimeter-level industrial real world insertion challenge, TAX-DPD outperforms a strong baseline in both unimodal and multi-modal settings, achieving success rates between 80 and 100\%.

% We present a hierarchical goal prediction framework that pairs a scene-conditioned Dense GMM for global placement with a local point-space diffuser that jointly denoises geometry and frame in normalized coordinates. TAX-DPD achieves state-of-the-art precision and coverage in simulation and outperforms a strong regression baseline on millimeter-level real-world insertion in both unimodal and multi-modal settings.

% \noindent\textbf{Limitation} Like most other works in object-centric goal prediction, our method requires accurate segmentations for the object $\mathcal{O}$. Additionally, we require correspondences between initial and goal configuration point clouds when collecting demonstration data. In the real world, this generally requires tracking, which can be very unreliable in the presence of occlusions. Notably, this is a result of our current feature processing pipeline, \textit{not} of the point cloud diffusion objective. In principle, our local configuration refinement architecture can be adapted to correspondence-free point cloud generation with minor modifications; we leave this for future work. \lyuxing{rewrite this}

\section{Acknowledgments} Supported by NSF CAREER Grant IIS-2046491. We also thank ABB Inc. for their support.

\bibliographystyle{IEEEtran}
\bibliography{ref}

@inproceedings{zhaolearning,
  title={Learning Fine-Grained Bimanual Manipulation with Low-Cost Hardware},
  author={Zhao, Tony Z and Kumar, Vikash and Levine, Sergey and Finn, Chelsea},
  booktitle={ICML Workshop on New Frontiers in Learning, Control, and Dynamical Systems}
}

@inproceedings{chi2024universal,
  title={Universal Manipulation Interface: In-The-Wild Robot Teaching Without In-The-Wild Robots},
  author={Chi, Cheng and Xu, Zhenjia and Pan, Chuer and Cousineau, Eric and Burchfiel, Benjamin and Feng, Siyuan and Tedrake, Russ and Song, Shuran},
  year={2024},
  organization={Robotics: Science and Systems}
}

@inproceedings{wutidybot++,
  title={TidyBot++: An Open-Source Holonomic Mobile Manipulator for Robot Learning},
  author={Wu, Jimmy and Chong, William and Holmberg, Robert and Prasad, Aaditya and Gao, Yihuai and Khatib, Oussama and Song, Shuran and Rusinkiewicz, Szymon and Bohg, Jeannette},
  booktitle={8th Annual Conference on Robot Learning}
}

@inproceedings{wu2024gello,
  title={Gello: A general, low-cost, and intuitive teleoperation framework for robot manipulators},
  author={Wu, Philipp and Shentu, Yide and Yi, Zhongke and Lin, Xingyu and Abbeel, Pieter},
  booktitle={2024 IEEE/RSJ International Conference on Intelligent Robots and Systems (IROS)},
  pages={12156--12163},
  year={2024},
  organization={IEEE}
}

@inproceedings{chengopen,
  title={Open-TeleVision: Teleoperation with Immersive Active Visual Feedback},
  author={Cheng, Xuxin and Li, Jialong and Yang, Shiqi and Yang, Ge and Wang, Xiaolong},
  booktitle={8th Annual Conference on Robot Learning}
}

@inproceedings{zhaoaloha2,
  title={ALOHA Unleashed: A Simple Recipe for Robot Dexterity},
  author={Zhao, Tony Z and Tompson, Jonathan and Driess, Danny and Florence, Pete and Ghasemipour, Seyed Kamyar Seyed and Finn, Chelsea and Wahid, Ayzaan},
  booktitle={8th Annual Conference on Robot Learning}
}

@article{black2024pi_0,
  title={$\pi_0$: A Vision-Language-Action Flow Model for General Robot Control},
  author={Black, Kevin and Brown, Noah and Driess, Danny and Esmail, Adnan and Equi, Michael and Finn, Chelsea and Fusai, Niccolo and Groom, Lachy and Hausman, Karol and Ichter, Brian and others},
  journal={arXiv preprint arXiv:2410.24164},
  year={2024}
}

@article{wang2025articubot,
  title={ArticuBot: Learning Universal Articulated Object Manipulation Policy via Large Scale Simulation},
  author={Wang, Yufei and Wang, Ziyu and Nakura, Mino and Bhowal, Pratik and Kuo, Chia-Liang and Chen, Yi-Ting and Erickson, Zackory and Held, David},
  journal={arXiv preprint arXiv:2503.03045},
  year={2025}
}

@inproceedings{chi2023diffusion,
  title={Diffusion Policy: Visuomotor Policy Learning via Action Diffusion},
  author={Chi, Cheng and Feng, Siyuan and Du, Yilun and Xu, Zhenjia and Cousineau, Eric and Burchfiel, Benjamin and Song, Shuran},
  booktitle={Robotics: Science and Systems},
  year={2023}
}

@inproceedings{ze2024dp3,
  title={3D Diffusion Policy: Generalizable Visuomotor Policy Learning via Simple 3D Representations},
  author={Ze, Yanjie and Zhang, Gu and Zhang, Kangning and Hu, Chenyuan and Wang, Muhan and Xu, Huazhe},
  booktitle={2nd Workshop on Dexterous Manipulation: Design, Perception and Control (RSS)}
}

@article{shafiullah2022behavior,
  title={Behavior transformers: Cloning $ k $ modes with one stone},
  author={Shafiullah, Nur Muhammad and Cui, Zichen and Altanzaya, Ariuntuya Arty and Pinto, Lerrel},
  journal={Advances in neural information processing systems},
  volume={35},
  pages={22955--22968},
  year={2022}
}

@inproceedings{lee2024behavior,
  title={Behavior Generation with Latent Actions},
  author={Lee, Seungjae and Wang, Yibin and Etukuru, Haritheja and ovKim, H Jin and Shafiullah, Nur Muhammad Mahi and Pinto, Lerrel},
  booktitle={International Conference on Machine Learning},
  pages={26991--27008},
  year={2024},
  organization={PMLR}
}

@inproceedings{zhen20243d,
  title={3D-VLA: A 3D Vision-Language-Action Generative World Model},
  author={Zhen, Haoyu and Qiu, Xiaowen and Chen, Peihao and Yang, Jincheng and Yan, Xin and Du, Yilun and Hong, Yining and Gan, Chuang},
  booktitle={International Conference on Machine Learning},
  pages={61229--61245},
  year={2024},
  organization={PMLR}
}

@inproceedings{cai2024tax3d,
  title={Non-rigid Relative Placement through 3D Dense Diffusion},
  author={Cai, Eric and Donca, Octavian and Eisner, Ben and Held, David},
  booktitle={Conference on Robot Learning (CoRL)},
  year={2024}
}

@inproceedings{pan2023tax,
  title={Tax-pose: Task-specific cross-pose estimation for robot manipulation},
  author={Pan, Chuer and Okorn, Brian and Zhang, Harry and Eisner, Ben and Held, David},
  booktitle={Conference on Robot Learning},
  pages={1783--1792},
  year={2023},
  organization={PMLR}
}

@inproceedings{huang2024imagination,
  title={IMAGINATION POLICY: Using Generative Point Cloud Models for Learning Manipulation Policies},
  author={Huang, Haojie and Schmeckpeper, Karl and Wang, Dian and Biza, Ondrej and Qian, Yaoyao and Liu, Haotian and Jia, Mingxi and Platt, Robert and Walters, Robin},
  year={2024},
  booktitle={Proceedings of the Conference on Robot Learning}
}

@inproceedings{simeonov2023shelving,
  title={Shelving, Stacking, Hanging: Relational Pose Diffusion for Multi-modal Rearrangement},
  author={Simeonov, Anthony and Goyal, Ankit and Manuelli, Lucas and Lin, Yen-Chen and Sarmiento, Alina and Garcia, Alberto Rodriguez and Agrawal, Pulkit and Fox, Dieter},
  booktitle={Conference on Robot Learning},
  pages={2030--2069},
  year={2023},
  organization={PMLR}
}

@article{zhao2025anyplace,
  title={AnyPlace: Learning Generalized Object Placement for Robot Manipulation},
  author={Zhao, Yuchi and Bogdanovic, Miroslav and Luo, Chengyuan and Tohme, Steven and Darvish, Kourosh and Aspuru-Guzik, Al{\'a}n and Shkurti, Florian and Garg, Animesh},
  journal={arXiv preprint arXiv:2502.04531},
  year={2025}
}

@inproceedings{liu2023structdiffusion,
  title={StructDiffusion: Language-Guided Creation of Physically-Valid Structures using Unseen Objects},
  author={Liu, Weiyu and Du, Yilun and Hermans, Tucker and Chernova, Sonia and Paxton, Chris},
  booktitle={Robotics: Science and Systems},
  year={2023}
}

@inproceedings{wang2024learning,
  title={Learning Distributional Demonstration Spaces for Task-Specific Cross-Pose Estimation},
  author={Wang, Jenny and Donca, Octavian and Held, David},
  booktitle={2024 IEEE International Conference on Robotics and Automation (ICRA)},
  pages={15054--15060},
  year={2024},
  organization={IEEE}
}

@inproceedings{eisnerdeep,
  title={Deep SE (3)-Equivariant Geometric Reasoning for Precise Placement Tasks},
  author={Eisner, Ben and Yang, Yi and Davchev, Todor and Vecerik, Mel and Scholz, Jonathan and Held, David},
  booktitle={The Twelfth International Conference on Learning Representations}
}

@inproceedings{simeonov2022neural,
  title={Neural descriptor fields: Se (3)-equivariant object representations for manipulation},
  author={Simeonov, Anthony and Du, Yilun and Tagliasacchi, Andrea and Tenenbaum, Joshua B and Rodriguez, Alberto and Agrawal, Pulkit and Sitzmann, Vincent},
  booktitle={2022 International Conference on Robotics and Automation (ICRA)},
  pages={6394--6400},
  year={2022},
  organization={IEEE}
}

@inproceedings{florence2018dense,
  title={Dense Object Nets: Learning Dense Visual Object Descriptors By and For Robotic Manipulation},
  author={Florence, Peter R and Manuelli, Lucas and Tedrake, Russ},
  booktitle={Conference on Robot Learning},
  pages={373--385},
  year={2018},
  organization={PMLR}
}

@inproceedings{chang2024dap,
  title={DAP: Diffusion-based Affordance Prediction for Multi-modality Storage},
  author={Chang, Haonan and Boyalakuntla, Kowndinya and Liu, Yuhan and Zhang, Xinyu and Schramm, Liam and Boularias, Abdeslam},
  booktitle={2024 IEEE/RSJ International Conference on Intelligent Robots and Systems (IROS)},
  pages={9476--9481},
  year={2024},
  organization={IEEE}
}

@article{ho2020denoising,
  title={Denoising diffusion probabilistic models},
  author={Ho, Jonathan and Jain, Ajay and Abbeel, Pieter},
  journal={Advances in neural information processing systems},
  volume={33},
  pages={6840--6851},
  year={2020}
}

@inproceedings{nichol2021improved,
  title={Improved denoising diffusion probabilistic models},
  author={Nichol, Alexander Quinn and Dhariwal, Prafulla},
  booktitle={International conference on machine learning},
  pages={8162--8171},
  year={2021},
  organization={PMLR}
}

@inproceedings{peebles2023scalable,
  title={Scalable diffusion models with transformers},
  author={Peebles, William and Xie, Saining},
  booktitle={Proceedings of the IEEE/CVF international conference on computer vision},
  pages={4195--4205},
  year={2023}
}

@article{brock2016generative,
  title={Generative and discriminative voxel modeling with convolutional neural networks},
  author={Brock, Andrew and Lim, Theodore and Ritchie, James M and Weston, Nick},
  journal={arXiv preprint arXiv:1608.04236},
  year={2016}
}

@inproceedings{kim2021setvae,
  title={Setvae: Learning hierarchical composition for generative modeling of set-structured data},
  author={Kim, Jinwoo and Yoo, Jaehoon and Lee, Juho and Hong, Seunghoon},
  booktitle={Proceedings of the IEEE/CVF Conference on Computer Vision and Pattern Recognition},
  pages={15059--15068},
  year={2021}
}

@inproceedings{achlioptas2018learning,
  title={Learning representations and generative models for 3d point clouds},
  author={Achlioptas, Panos and Diamanti, Olga and Mitliagkas, Ioannis and Guibas, Leonidas},
  booktitle={International conference on machine learning},
  pages={40--49},
  year={2018},
  organization={PMLR}
}

@inproceedings{yang2021cpcgan,
  title={Cpcgan: A controllable 3d point cloud generative adversarial network with semantic label generating},
  author={Yang, Ximing and Wu, Yuan and Zhang, Kaiyi and Jin, Cheng},
  booktitle={Proceedings of the AAAI conference on artificial intelligence},
  volume={35},
  number={4},
  pages={3154--3162},
  year={2021}
}

@inproceedings{shu20193d,
  title={3d point cloud generative adversarial network based on tree structured graph convolutions},
  author={Shu, Dong Wook and Park, Sung Woo and Kwon, Junseok},
  booktitle={Proceedings of the IEEE/CVF international conference on computer vision},
  pages={3859--3868},
  year={2019}
}

@inproceedings{luo2021diffusion,
  title={Diffusion probabilistic models for 3d point cloud generation},
  author={Luo, Shitong and Hu, Wei},
  booktitle={Proceedings of the IEEE/CVF conference on computer vision and pattern recognition},
  pages={2837--2845},
  year={2021}
}

@inproceedings{zyrianov2022learning,
  title={Learning to generate realistic lidar point clouds},
  author={Zyrianov, Vlas and Zhu, Xiyue and Wang, Shenlong},
  booktitle={European Conference on Computer Vision},
  pages={17--35},
  year={2022},
  organization={Springer}
}

@article{dahnert2024coherent,
  title={Coherent 3D Scene Diffusion From a Single RGB Image},
  author={Dahnert, Manuel and Dai, Angela and M{\"u}ller, Norman and Nie{\ss}ner, Matthias},
  journal={Advances in Neural Information Processing Systems},
  volume={37},
  pages={23435--23463},
  year={2024}
}

@article{nichol2022point,
  title={Point-e: A system for generating 3d point clouds from complex prompts},
  author={Nichol, Alex and Jun, Heewoo and Dhariwal, Prafulla and Mishkin, Pamela and Chen, Mark},
  journal={arXiv preprint arXiv:2212.08751},
  year={2022}
}

@article{mo2023dit,
  title={Dit-3d: Exploring plain diffusion transformers for 3d shape generation},
  author={Mo, Shentong and Xie, Enze and Chu, Ruihang and Hong, Lanqing and Niessner, Matthias and Li, Zhenguo},
  journal={Advances in neural information processing systems},
  volume={36},
  pages={67960--67971},
  year={2023}
}

@inproceedings{mo2024fast,
  title={Fast training of diffusion transformer with extreme masking for 3d point clouds generation},
  author={Mo, Shentong and Xie, Enze and Wu, Yue and Chen, Junsong and Nie{\ss}ner, Matthias and Li, Zhenguo},
  booktitle={European Conference on Computer Vision},
  pages={354--370},
  year={2024},
  organization={Springer}
}

@article{huang2022city3d,
  title={City3D: Large-scale building reconstruction from airborne LiDAR point clouds},
  author={Huang, Jin and Stoter, Jantien and Peters, Ravi and Nan, Liangliang},
  journal={Remote Sensing},
  volume={14},
  number={9},
  pages={2254},
  year={2022},
  publisher={MDPI}
}

@inproceedings{sanghi2022clip,
  title={Clip-forge: Towards zero-shot text-to-shape generation},
  author={Sanghi, Aditya and Chu, Hang and Lambourne, Joseph G and Wang, Ye and Cheng, Chin-Yi and Fumero, Marco and Malekshan, Kamal Rahimi},
  booktitle={Proceedings of the IEEE/CVF Conference on Computer Vision and Pattern Recognition},
  pages={18603--18613},
  year={2022}
}

@article{lee2023diffusion,
  title={Diffusion probabilistic models for scene-scale 3d categorical data},
  author={Lee, Jumin and Im, Woobin and Lee, Sebin and Yoon, Sung-Eui},
  journal={arXiv preprint arXiv:2301.00527},
  year={2023}
}

@inproceedings{ran2024towards,
  title={Towards realistic scene generation with lidar diffusion models},
  author={Ran, Haoxi and Guizilini, Vitor and Wang, Yue},
  booktitle={Proceedings of the IEEE/CVF Conference on Computer Vision and Pattern Recognition},
  pages={14738--14748},
  year={2024}
}

@InProceedings{Zhou_2021_ICCV,
    author    = {Zhou, Linqi and Du, Yilun and Wu, Jiajun},
    title     = {3D Shape Generation and Completion Through Point-Voxel Diffusion},
    booktitle = {Proceedings of the IEEE/CVF International Conference on Computer Vision (ICCV)},
    month     = {October},
    year      = {2021},
    pages     = {5826-5835}
}

@article{qi2017pointnet++,
  title={Pointnet++: Deep hierarchical feature learning on point sets in a metric space},
  author={Qi, Charles Ruizhongtai and Yi, Li and Su, Hao and Guibas, Leonidas J},
  journal={Advances in neural information processing systems},
  volume={30},
  year={2017}
}

@inproceedings{antonova2021dynamic,
  title={Dynamic environments with deformable objects},
  author={Antonova, Rika and Shi, Peiyang and Yin, Hang and Weng, Zehang and Jensfelt, Danica Kragic},
  booktitle={Thirty-fifth Conference on Neural Information Processing Systems Datasets and Benchmarks Track (Round 2)},
  year={2021}
}

@misc{bullet,
    title={PyBullet, a Python module for physics simulation for games, robotics, and machine learning},
    author={Erwin Coumans and Yunfei Bai},
    url={http://pybullet.org},
    year={2016-2020}
}

@inproceedings{xian2023chaineddiffuser,
  title={Chaineddiffuser: Unifying trajectory diffusion and keypose prediction for robotic manipulation},
  author={Xian, Zhou and Gkanatsios, Nikolaos},
  booktitle={Conference on Robot Learning/Proceedings of Machine Learning Research},
  year={2023},
  organization={Proceedings of Machine Learning Research}
}

@article{nist_assembly_2018,
	title = {Assembly {Performance} {Metrics} and {Test} {Methods}},
	url = {https://www.nist.gov/el/intelligent-systems-division-73500/robotic-grasping-and-manipulation-assembly/assembly},
	language = {en},
	urldate = {2024-10-16},
	journal = {NIST},
	month = may,
	year = {2018},
}

@inproceedings{xu2023iterative,
  title={Iterative geometry encoding volume for stereo matching},
  author={Xu, Gangwei and Wang, Xianqi and Ding, Xiaohuan and Yang, Xin},
  booktitle={Proceedings of the IEEE/CVF conference on computer vision and pattern recognition},
  pages={21919--21928},
  year={2023}
}

@inproceedings{zhou2019continuity,
  title={On the continuity of rotation representations in neural networks},
  author={Zhou, Yi and Barnes, Connelly and Lu, Jingwan and Yang, Jimei and Li, Hao},
  booktitle={Proceedings of the IEEE/CVF conference on computer vision and pattern recognition},
  pages={5745--5753},
  year={2019}
}


\clearpage
\appendices
\renewcommand{\thesection}{S\arabic{section}}
\setcounter{section}{0}

\begin{center}
    {\LARGE \textbf{Supplementary Material}}\\[1em]
\end{center}

\section{Extension to Non-Rigid Placement Tasks}
\label{sec:nonrigid_experiments}
\subsection{Experimental Setup}
\begin{figure*}[h]
    \centering
    \includegraphics[width=\linewidth]{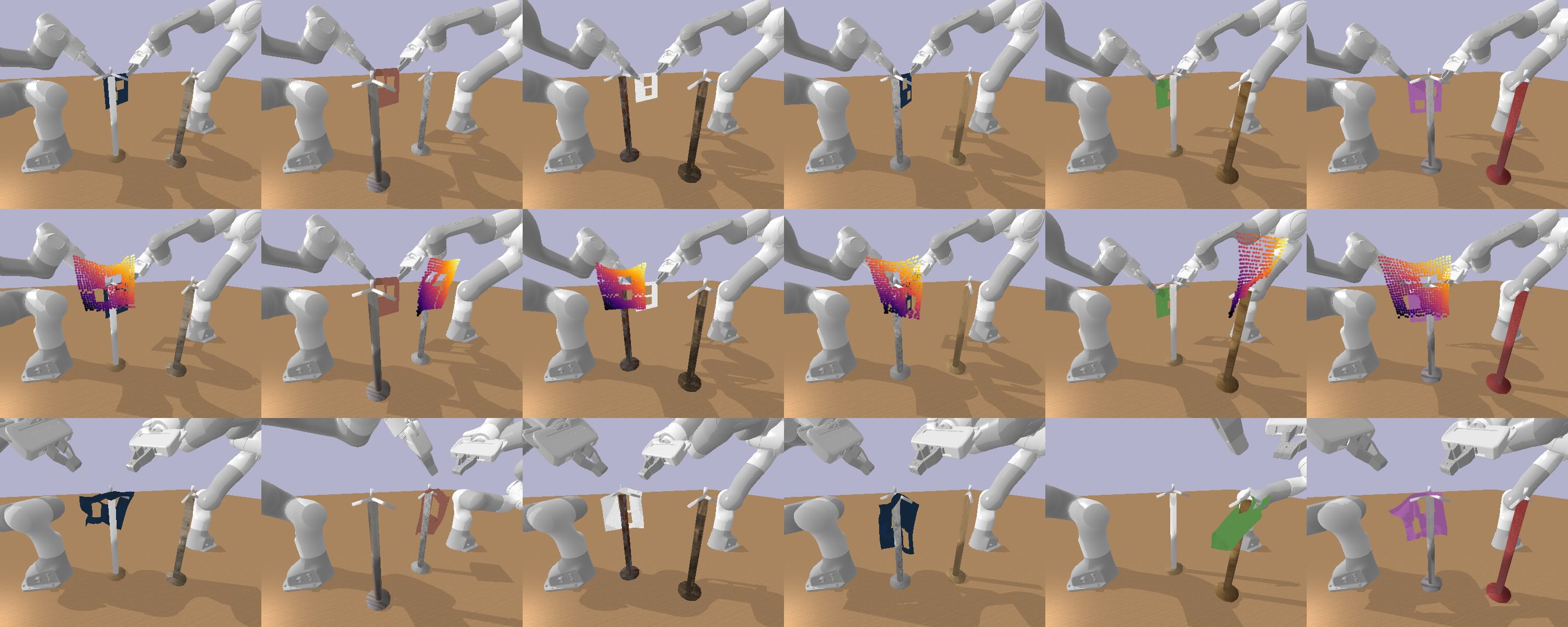}
    \caption{\textbf{DEDO Task Environment.} (\textit{Top}) A visualization of the variation in configuration in the \texttt{HangProcCloth-DH} task. (\textit{Middle}) Our method generalizes well to diverse cloth geometries. (\textit{Bottom}) Successful placements with a goal-conditioned placement policy.}
    \label{fig:app_dedo_samples}
\end{figure*}

Since our point cloud–based formulation for goal prediction does not assume object rigidity, our method can be naturally applied to deformable objects without requiring any architecture modifications. We validate this capability on a cloth hanging task from Dynamic Environments with Deformable Objects (DEDO)~\cite{antonova2021dynamic} built on PyBullet physics engine~\cite{bullet}. Specifically, we conduct experiments on a modified multi-hanger implementation of the \texttt{HangProcCloth} task, in which a procedurally generated cloth with two holes must be hung on one of the two hangers in the scene. The task explores multi-modal non-rigid placements under significant variation in object shape and scene layout (see Figure~\ref{fig:app_dedo_samples}). 

%Since our point cloud–based formulation for goal prediction does not assume object rigidity, our method can be naturally applied to deformable objects without requiring any architectural modifications. We validate this capability on cloth-hanging tasks from Dynamic Environments with Deformable Objects (DEDO)~\cite{antonova2021dynamic} built on the PyBullet physics engine~\cite{bullet}. Specifically, we conduct experiments on a modified multi-hanger implementation of the HangProcCloth task, in which a procedurally generated cloth must be hung on one of the hangers in the scene. We evaluate on two variations: SH (single-hole), in which the cloth contains one hole, and DH (double-hole), in which the cloth contains two holes. These tasks explore multi-modal non-rigid placement under significant variation in object shape and scene layout (see Figure~\ref{fig:app_dedo_samples}). To further increase task complexity and realism, we introduce several modifications to the environment: the floating grippers for cloth control are replaced with a bimanual Franka arm setup, the initial pose of the cloth is randomized at the start of each rollout, and the scene is configured with multiple hangers to induce multi-modality. Together, these modifications highlight the challenges of extending object-centric goal prediction to deformable settings, where open-loop execution strategies and pre-defined motions are insufficient to guarantee success (see Figure~\ref{fig:app_dedo_failures}).

\subsection{Goal-conditioned Policy Learning}
\label{sec:goal_conditioned_policy}
In RPDiff, the predicted goal configurations are executed via motion planning and task-specific primitives, or in other words using manually defined pre-placement offset poses based on task-dependent heuristics to avoid collisions during execution. However, in deformable settings, such heuristics and motion planning cannot guarantee success, since—unlike rigid objects where aligning a key part suffices—the object’s shape continuously changes under manipulation, making fixed pre-placement offsets unreliable.

Alternatively, recent work~\cite{wang2025articubot,xian2023chaineddiffuser} has demonstrated success with hierarchical policy learning with explicitly 3D high-level goals. ArticuBot~\cite{wang2025articubot}, for example, predicts an action-centric high-level goal (represented directly in 3D as a set of gripper points) to condition a lower-level policy. We follow the similar ideology to train a placement policy conditioned on our predicted goal configuration $\hat{P}^*_{\mathcal{O}}$ to accomplish the challenging deformable cloth hanging task. Specifically, we use an exceedingly simple framework in which we concatenate $\hat{P}^*_{\mathcal{O}}$ to the point cloud observation input for 3D Diffusion Policy (DP3)~\cite{ze2024dp3}. To avoid iteratively denoising $\hat{P}^*_{\mathcal{O}}$ during training, which can be slow, we condition on the ground-truth $P^*_{\mathcal{O}}$ when training the low-level policy, and condition on the predicted goal configuration $\hat{P}^*_{\mathcal{O}}$ during task evaluation.

\subsection{Demonstration Generation }
To collect demonstrations in the \texttt{HangProcCloth} environment, we design a task-specific expert policy that applies end-effector position control with soft constraints to (i) smoothly orient the cloth toward the hanger, (ii) maintain the gripper distance approximately equal to the cloth width, and (iii) keep the grippers at similar heights along the $z$-axis. Each demonstration provides data for two complementary objectives: for goal prediction, we collect triplets $(P_\mathcal{O}, P_\mathcal{S}, P_\mathcal{O}^*)$ where $P_\mathcal{O} \in \mathbb{R}^{N_\mathcal{O}\times 3}$ is the segmented object point cloud, $P_\mathcal{S} \in \mathbb{R}^{N_\mathcal{S}\times 3}$ is the scene point cloud, and $P_\mathcal{O}^* \in \mathbb{R}^{N_\mathcal{O}\times 3}$ is the ground-truth goal configuration, as we defined earlier in the problem statement. For policy learning, we collect tuples $(\{o_t\}_{t=1}^T, \{a_t\}_{t=1}^T, P_\mathcal{O}^*)$ where $o_t \in \mathcal{O}$ are point cloud observations, $a_t \in \mathcal{A}$ are end-effector actions, and $P_\mathcal{O}^*$ is the conditioning goal. These demonstrations allow us to train our method for predicting goal placement configuration for the cloth as well as a placement policy (goal conditioned 3D Diffusion Policy) to execute the prediction. 

\subsection{Evaluation and Metrics}
For the cloth hanging task, we evaluate success rates over 100 trials, where in each trial the initial pose of the cloth and the hangers are both randomized, and the cloth's geometry is also randomized by varying the holes' positions (unseen during training). Success is determined by evaluating the final simulator state, whether a hole of the cloth passes fully through the rod of one of the hangers in the scene after the trained placement policy executes the predicted placement configuration. Additionally, we quantify multimodal capture and placement precision via reporting \textbf{Coverage RMSE}, defined as the minimum point-wise root-mean-square error (RMSE) between a ground-truth goal configuration $P_\mathcal{O}^*$ and a set of $K$ predicted samples $\{\hat{P}_{\mathcal{O}}^{*(k)}\}_{k=1}^{K}$, and \textbf{Precision RMSE}, defined as the minimum RMSE between a single predicted sample $\hat{P}_\mathcal{O}^*$ and a set of $M$ ground-truth goal configurations $\{P_{\mathcal{O}}^{*(m)}\}_{m=1}^{M}$.

\subsection{Baselines}
We train and compare against \textbf{TAX3D}~\cite{cai2024tax3d}, a point cloud diffusion approach for non-rigid, multi-modal placement. To ensure a fair comparison, during success rate evaluation we execute the predicted placements from TAX3D using the same learned placement policy described in Sec~\ref{sec:goal_conditioned_policy}.

\begin{table}[t]
    \centering
    \caption{Ablations and Task Success Rates on modified DEDO \texttt{HangProcCloth} task.}
    \label{tab:results_table_sim_nonrigid}
    \begin{tabular}{l|ccc}
    \toprule
        & SR & Cov. RMSE & Prec. RMSE\\
         \midrule
        TAX3D~\cite{cai2024tax3d} & 0.50 & 0.87 & 1.34 \\
        \midrule
        \textit{Ours} w/o disentangled point diffusion & 0.73 & 0.82 & 1.16\\
        \textit{Ours} w/ MLP encodings & 0.76 & 0.96 & 1.04 \\
        \textit{Ours} w/o GMM & 0.69 & 0.88 & 0.96 \\
        \textit{Ours} w/o recon. embedding & 0.75 & 0.82 & 0.81 \\
        \textit{Ours} w/o rot. noise & 0.64 & 0.64 & 0.95 \\
        \textit{Ours} w/o deform. embedding & 0.71 & 0.96 & 0.95 \\
        \midrule
        \textit{Ours} & \textbf{0.78} & \textbf{0.50} & \textbf{0.58} \\
        \bottomrule
    \end{tabular}
\end{table}

\subsection{Results and Ablations}
As shown in Table~\ref{tab:results_table_sim_nonrigid}, our method achieves substantially better performance than the TAX3D baseline on the \texttt{HangProcCloth} task, improving success rate by a large margin (0.50 $\rightarrow$ 0.78) while also drastically reducing coverage RMSE (0.87 $\rightarrow$ 0.50) and precision RMSE (1.34 $\rightarrow$ 0.58). Removing \emph{disentangled point diffusion} reduces success rate by 5 percents, while coverage and precision RMSE increase by 0.32 and 0.58, respectively. This highlights the critical role of explicitly separating frame and shape denoising rather than conflating them in a single point-space process. Similarly, removing the scene-conditioned \emph{Dense GMM} for global initialization lowers success rate by 9 points and degrades both coverage and precision RMSE by 0.38 each, confirming its importance in capturing multimodality at the global scene level. All the remaining components also contribute meaningfully to performance. In particular, ablating the deformation embedding reduces success by 7 points and worsens coverage and precision RMSE by 0.46 and 0.37, a much larger relative effect than observed in rigid-object experiments. This underscores the value of explicitly encoding deformation-specific cues in non-rigid placement tasks, where object geometry evolves dynamically under manipulation.

\section{Training Details}
\subsection{Training Pre-processing}
\subsubsection{Dataset Scaling}
As one of the motivating issues of our method, diffusing point clouds in the object placement setting is less feasible due to the scale mismatch between the object and the scene. To address this and ensure consistency across tasks, we standardize the input space by adaptively scaling both the scene and object point clouds based on task-specific statistics. Specifically, we compute normalization factors from the empirical distribution of point cloud scales for each task, allowing our method to operate in a unified and numerically stable space that maximizes the effectiveness of goal configuration prediction. The scale normalization of point clouds is done automatically for each batch. 

\subsubsection{Point Cloud Downsampling}
For training, both the object and scene point clouds are downsampled using furthest point sampling. Depending on the scene complexity, we vary the number of sampled points to ensure minimal geometric information loss, preserving the structural features necessary for inferring the goal object configuration. We document the number of points for the objects and the scenes for each task in Table~\ref{tab:downsampling}.

\subsubsection{Augmentation}
We additionally augment the scene, initial object, and goal configuration point clouds with the same rotation, which is uniformly sampled from $[0, 2\pi]$ about the $z$-axis.

\subsection{Hyper-parameters}
We provide the hyper-parameters used for training and model configuration in Table~\ref{tab:hyperparams}. These include both optimization settings (e.g., batch size, learning rate, weight decay) and architectural choices for the diffusion transformer backbone.
\begin{table}[b!]
    \centering
    \caption{Training and Model Hyper-parameters.}
    \label{tab:hyperparams}
    \begin{tabular}{ll}
        \toprule
        \textbf{Parameter} & \textbf{Value} \\
        \midrule
        Batch size & 16 \\
        Learning rate & 1e-4 \\
        Learning rate warmup steps & 100 \\
        Weight decay & 1e-5\\
        Epochs & 20,000\\
        \midrule
        Number of DiT blocks & 5 \\
        Number of heads per block & 4 \\
        Hidden size per block & 128 \\
        Diffusion steps & 100 \\
    \bottomrule
    \end{tabular}
    \vspace{0.5em}
\label{tab:training_hparams}
\end{table}

\begin{table}[t]
    \centering
    \caption{Point Cloud Downsampling per Task.}
    \label{tab:downsampling}
    \begin{tabular}{c|l}
         \texttt{RPDiff-Mug/EasyRack} & 
         \makecell[l]{number of $P_\mathcal{O}$ points = 512 \\ number of $P_\mathcal{S}$ points = 512} \\ \midrule
         
         \texttt{RPDiff-Mug/MedRack} & 
         \makecell[l]{number of $P_\mathcal{O}$ points = 512 \\ number of $P_\mathcal{S}$ points = 512} \\ \midrule
         
         \texttt{RPDiff-Mug/Multi-MedRack} & 
         \makecell[l]{number of $P_\mathcal{O}$ points = 512 \\ number of $P_\mathcal{S}$ points = 1024} \\ \midrule
         
         \texttt{RPDiff-Book/Shelf} & 
         \makecell[l]{number of $P_\mathcal{O}$ points = 512 \\ number of $P_\mathcal{S}$ points = 1024} \\ \midrule
         
         \texttt{RPDiff-Can/Cabinet} & 
         \makecell[l]{number of $P_\mathcal{O}$ points = 256 \\ number of $P_\mathcal{S}$ points = 1024} \\ \midrule

         \texttt{DEDO-HangProcCloth} & 
         \makecell[l]{number of $P_\mathcal{O}$ points = 512 \\ number of $P_\mathcal{S}$ points = 512}
    \end{tabular}
\end{table}

\section{Method Details}
%\lyuxing{prob remake our modified DiT architecture diagram?}
\subsection{Global Placement Initialization - Architecture}
\label{sec:app_global_place}
To predict the Dense GMM from which we sample $\hat{g}$, we use a simplified version of the point cloud diffusion architecture for local configuration refinement. Namely, we encode the object point cloud $P_{\mathcal{O}}$ and the scene point cloud $P_\mathcal{S}$ separately with PointNet++~\cite{qi2017pointnet++}, before passing the object and scene tokens into a point cloud transformer. This transformer has the same underlying architecture as the modified DiT in local configuration refinement, with two exceptions: 1) the scene tokens cross-attend to the object tokens (rather than the object tokens cross-attending to the scene tokens), and 2) the timestep $t$ is set to 0, since gloal placement initialization does not do any iterative denoising. The final scene tokens are then decoded by a pointwise MLP into per-point weights $\{w_i\}_{i=1}^{N_\mathcal{S}}$ and per-point residuals $\{r_i\}_{i=1}^{N_\mathcal{S}}$, which parameterize our proposed Dense GMM..

\subsection{Local Configuration Refinement - Rotation Noise}
% \subsubsection{Rotation Noise}
\label{sec:app_local_rotation}
During the forward diffusion process applied to the shape point cloud $\phi_0 \in \mathbb{R}^{3 \times N}$, we inject both standard Gaussian noise and an additional rigid-body rotation noise. The perturbed shape at diffusion timestep $t$ is defined as
\[
\phi_t = \sqrt{\bar{\alpha}_t} \, \phi_0 + \sqrt{1 - \bar{\alpha}_t} \, (\boldsymbol{\epsilon}_{\phi} + \boldsymbol{\epsilon}_{\mathrm{rot}})
\]
where $\bar{\alpha}_t \in (0, 1)$ denotes the cumulative product of noise scheduling coefficients, $\boldsymbol{\epsilon}_{\phi} \sim \mathcal{N}(0, \mathbf{I}) \in \mathbb{R}^{3 \times N}$ is the per-point i.i.d. Gaussian noise for the object shape forward corruption processes, and $\boldsymbol{\epsilon}_{\mathrm{rot}} \in \mathbb{R}^{3 \times N}$ represents the rigid-body rotation noise.

To compute $\boldsymbol{\epsilon}_{\mathrm{rot}}$, we first sample a random rotation axis $\mathbf{u} \in \mathbb{R}^3$ uniformly on the unit sphere, i.e.,
\[
\mathbf{u} \sim \mathcal{U}(\mathbb{S}^2), \quad \text{s.t.} \: \|\mathbf{u}\|_2 = 1
\]
and a rotation angle $\theta \in \mathbb{R}$ from a zero-mean Gaussian distribution,
\[
\theta \sim \mathcal{N}(0, \sigma_{\mathrm{rot}}^2)
\]
We empirically set $\sigma_{\mathrm{rot}} = \frac{\pi}{4}$ (i.e., 45 degrees). Given $(\mathbf{u}, \theta)$, we construct the corresponding rotation matrix $R \in \mathrm{SO}(3)$ using Rodrigues' rotation formula:
\[
R = \mathbf{I}_3 + \sin(\theta)[\mathbf{u}]_\times + (1 - \cos(\theta))[\mathbf{u}]_\times^2
\]
where $[\mathbf{u}]_\times \in \mathbb{R}^{3 \times 3}$ is the skew-symmetric matrix of $\mathbf{u}$:
\[
[\mathbf{u}]_\times =
\begin{bmatrix}
0 & -u_3 & u_2 \\
u_3 & 0 & -u_1 \\
-u_2 & u_1 & 0
\end{bmatrix}
\]
We then define the rotation noise as the deviation between the rotated and original shape:
\[
\boldsymbol{\epsilon}_{\mathrm{rot}} = R \phi_0 - \phi_0
\]
This term is added to the standard diffusion noise to encourage the model to denoise both pointwise and global rotational perturbations. We empirically found this modification help the denoiser to better model pose transformations of the placement objects.

\subsection{Local Configuration Refinement - Token Mixing}
\label{sec:app_local_token_mixing}

After point cloud encoding in local configuration refinement, we obtain reconstruction embeddings $\{f_i\}_{i=1}^{N_\mathcal{O} + N_\mathcal{S}}$, object embeddings $\{o_j\}_{j=1}^{N_\mathcal{O}}$, and deformation embeddings $\{d_k\}_{k=1}^{N_\mathcal{O}}$. In particular, the reconstruction embeddings can be split into object-specific reconstruction embeddings $\{f^\mathcal{O}_i\}_{i=1}^{N_\mathcal{O}}$ and scene-specific reconstruction embeddings $\{f^\mathcal{S}_i\}_{i=1}^{N_\mathcal{S}}$. With $d$ denoting the dimension size of these embeddings, we concatenate the three sets of object-specific embeddings into object tokens $\mathcal{T}^\mathcal{O} = \{\begin{bmatrix}
    f_i^{\mathcal{O}} & o_i & d_i
\end{bmatrix}\}_{i=1}^{N_\mathcal{O}}$ such that $\mathcal{T}^\mathcal{O} \in \mathbb{R}^{N_\mathcal{O}\times 3d}$, and directly use the scene-specific reconstruction embeddings as scene tokens $\mathcal{T}^\mathcal{S} = \{f_i^{\mathcal{S}}\}_{i=1}^{N_\mathcal{S}}$ such that $\mathcal{T}^\mathcal{S} \in \mathbb{R}^{N_\mathcal{S}\times d}$.

Before passing these tokens into the modified Point Cloud Diffusion Transformer, we project the object tokens $\mathcal{T}^{\mathcal{O}}$ from a dimension size of $3d$ down to a dimension size of $d$ with a pointwise MLP to match the scene tokens $\mathcal{T}^{\mathcal{S}}$. We refer to this step as ``token mixing."

\subsection{Local Configuration Refinement - SE(3) Diffusion Ablation}
\label{sec:app_se_diffusion}
To adapt our local refinement stage to instead perform diffusion directly on the SE(3) manifold, we replace the object shape diffusion with a denoising process over the object's orientation, while maintaining the rest of the architecture of local configuration refinement components unchanged. Following prior work, we represent the orientation in a 6D continuous rotation format~\cite{zhou2019continuity}, which allows for the application of a standard Gaussian-noise-based DDPM process in this Euclidean space. Specifically, let $\hat{\mathbf{R}}_{t-1}\!\in\!\mathrm{SO}(3)$ and $\hat{\mathbf{t}}_{t-1}\!\in\!\mathbb{R}^3$ denote the denoised orientation and translation at step $t\!-\!1$ in the local frame $\hat{g}$. 
The denoiser predicts $(\hat{\mathbf{R}}_{t-1},\,\hat{\mathbf{t}}_{t-1})$ from the current estimate at step $t$, and we compose
\[
\hat{\mathbf{T}}_{t-1}
=
\begin{bmatrix}
\hat{\mathbf{R}}_{t-1} & \hat{\mathbf{t}}_{t-1}\\[2pt]
\mathbf{0}^\top & 1
\end{bmatrix}\in\mathrm{SE}(3)
\]
Applying this transform to the initial object point cloud expressed in the $\hat{g}$-frame, ${}^{\hat{g}}\!P_{\mathcal{O}}$, yields the denoised goal at step $t\!-\!1$:
\[
{}^{\hat{g}}\hat{P}^{*}_{\mathcal{O},\,t-1}
= 
\hat{\mathbf{T}}_{t-1}\,{}^{\hat{g}}\!P_{\mathcal{O}}
\]
with all quantities defined in the local placement frame $\hat{g}$.

\begin{figure}[b]
    \centering
    \includegraphics[width=\columnwidth]{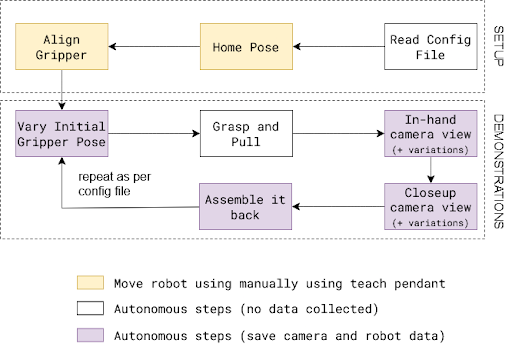}
    \caption{Data collection process for the vision-based insertion skill. The number of cycles, pose variation ranges, and setup parameters are defined using a YAML configuration file.}
    \label{fig:vision-teach}
\end{figure}

\begin{table*}[t]
    \centering
    \caption{Additional Real-World Unimodal Insertion Tasks Results.}
    {\scriptsize
    \begin{tabular}{c|c|cc|cc|cc|} 
         & & \multicolumn{2}{c}{Waterproof} 
         & \multicolumn{2}{c}{DSUB-25}
         & \multicolumn{2}{c}{SSD}\\
         \midrule
         & & TAX-Pose & \textit{Ours} 
         & TAX-Pose & \textit{Ours}  
         & TAX-Pose & \textit{Ours}  
         \\
         \midrule
            \multirow{2}{*}{Success} & Trans. Err. (mm) & 0.99 & \textbf{0.72} & 0.75 & \textbf{0.53} & - & \textbf{1.72} \\
            & Rot. Err. ($^\circ$) & 1.54 & \textbf{1.18} & 2.90 & \textbf{1.17} & - & \textbf{2.01} \\
        \midrule
            \multirow{2}{*}{Failure} & Trans. Err. (mm) & 1.35 \textcolor{red}{(+0.36)} & \textbf{-} \textcolor{red}{(-)} & \textbf{1.65} \textcolor{red}{(+0.90)} & 1.96 \textcolor{red}{(+1.43)} & 16.18 \textcolor{red}{(-)} & \textbf{6.94} \textcolor{red}{(+5.22)} \\
            & Rot. Err. ($^\circ$) & 2.26 \textcolor{red}{(+0.72)} & \textbf{-} \textcolor{red}{(-)} & 4.20 \textcolor{red}{(+1.30)} & \textbf{1.73} \textcolor{red}{(+0.56)} & 13.81 \textcolor{red}{(-)} & \textbf{8.75} \textcolor{red}{(+6.74)} \\
    \end{tabular}
    }
    \label{tab:results_table_real_additional}
\end{table*}

\begin{figure*}[t]
    \centering
    \begin{minipage}[t]{0.5\textwidth}
        \centering
        \includegraphics[width=\textwidth]{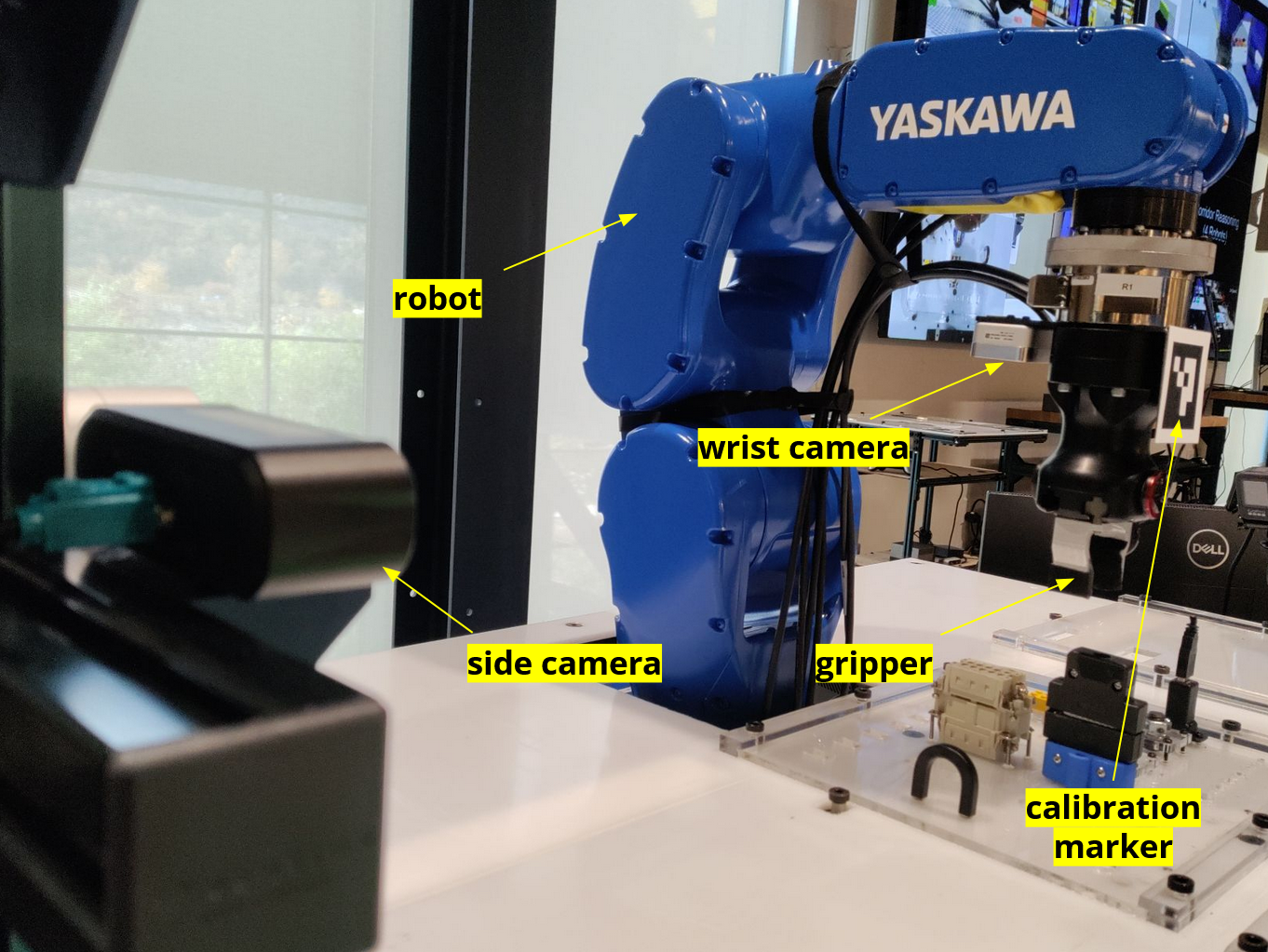}
        \caption{The vision-based insertion requires two cameras, one on the end effector (wrist camera, Intel D405), and the other fixed to ground (side camera, ZEDX-Mini). The setup shows Yaskawa GP4 robot, and Robotiq HandE gripper.}
        \label{fig:vision_setup}
    \end{minipage}
    \hspace{0.05\textwidth}
    \begin{minipage}[t]{0.43\textwidth}
        \centering
        \includegraphics[width=\textwidth]{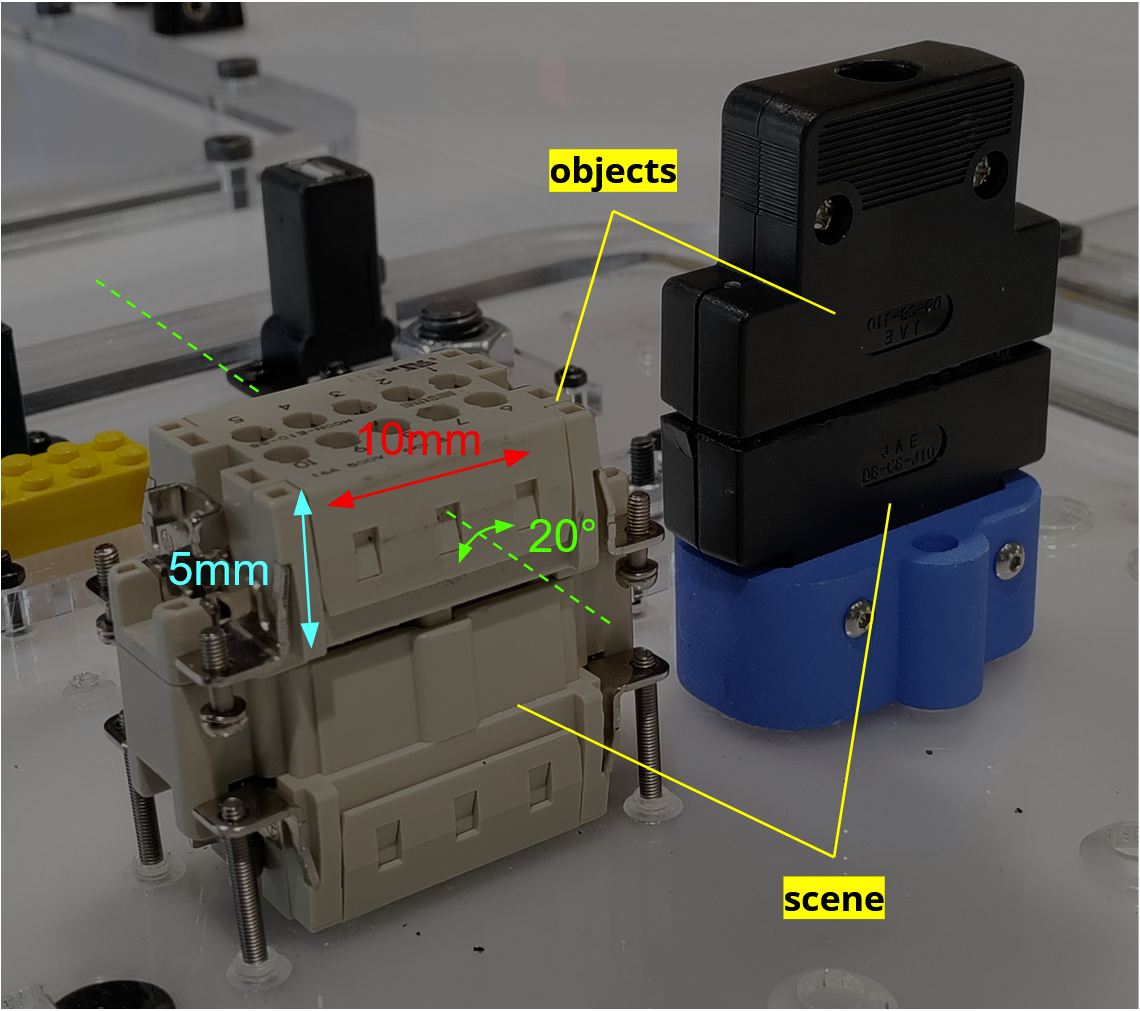}
        \caption{The conditioning scene is the receptacle or socket, and the object is the plug or connector. Random variations of \textcolor{green}{$20^\circ$ y-axis rotation}, \textcolor{red}{10mm x-axis translation}, and \textcolor{lightblue}{5mm x-axis translation} are applied to the objects.}  
        \label{fig:insertion_illustration}
    \end{minipage}
\end{figure*}

\subsection{RANSAC}
Given the sampled global placement reference frame $\hat{g}$ and the denoised local placement configuration ${^{\hat{g}}\hat{P}^*_\mathcal{O}}$, we can compute the global predicted goal placement point cloud as $\hat{P}^*_\mathcal{O} = {^{\hat{g}}\hat{P}^*_\mathcal{O}} + \hat{g}$. %Since our model design embeds per-point correspondence as an inductive bias, 
In our model design, each input point $p_i \in P_\mathcal{O}$ directly corresponds to a predicted goal point $\hat{p}^*_i \in \hat{P}^*_\mathcal{O}$.  

For some downstream applications, we may need to predict the final SE(3) pose of the target object.
However, since our diffusion process denoises points independently, the predicted goal set may contain local inconsistencies and outliers that violate rigid-body constraints. To address this, we employ a RANSAC-SVD approach as a  post-processing step that projects the noisy correspondences onto a single rigid-body transform. Specifically, RANSAC iteratively:
\begin{enumerate}
    \item samples a set of three correspondences $\{(p_i, \hat{p}^*_i)\}$, 
    \item estimates a candidate transformation $T$ from these correspondences, 
    \item evaluates inlier support by counting correspondences that satisfy $\|T p_j - \hat{p}^*_j\|_2 < \tau$, where $\tau$ is a distance threshold.
\end{enumerate}
After $N$ iterations, we select the transform with the largest inlier set, then re-estimate the final SE(3) transform using SVD over all inliers.

\section{Real World Experiments Details}
\subsection{Setup Descriptions}
\subsubsection{Hardware}
The hardware setup is shown in Figure~\ref{fig:vision_setup}. The robot uses a 6-DOF arm and a gripper to manipulate the objects. To get visual input for insertion pose estimation, our setup contains two cameras: one at the end effector (wrist camera, Intel D405), and the other fixed to the table on the side of the workspace (side camera,  ZEDX-Mini)). The objects chosen to train and perform insertion tasks are connectors from the NIST Assembly Task Board 1 \cite{nist_assembly_2018}. Specifically, we perform experiments with the ``Waterproof'' connector, the ``DSUB-25'' connector, and the ``SSD'' connector. Each connector has two parts: the ``object" $\mathcal{O}$ (the connector) and the ``scene" $\mathcal{S}$ (the socket). The scene $\mathcal{S}$ is fixed in place on the table surface while the object $\mathcal{O}$ is grasped by the robot gripper. The robot aims to infer the correct placement pose to insert the object into the scene successfully. 

\subsubsection{Capturing Point Cloud}
We use images from the stereo cameras and the deep stereo method IGEV \cite{xu2023iterative} to provide depth estimation and therefore scene point clouds. When capturing the plug part of the connector (i.e. the object $\mathcal{O}$), which is movable, the robot grasping the object $\mathcal{O}$ moves to an object-capturing pose such that the side camera, fixed to the ground, can capture the bottom surface of the connector, which is the contact interface for insertion. When capturing the receptacle part of the connector (i.e. the scene $\mathcal{S}$), which is fixed, the robot then moves to a scene-capturing pose such that the in-hand camera can capture a close-up view of the receptacle’s opening surface, which serves as the insertion target for the plug. Both the object-capturing and scene-capturing poses are tuned such that the target objects can get close to the cameras to capture point clouds of a desired fidelity. Once the point clouds are captured, we further apply a heuristic 3D bounding box, manually defined based on the object geometries, to crop or segment the object and scene point clouds from the full observation.

\subsubsection{Generating Demonstration Data.}
Data is collected by making the robot perform the insertion task multiple times with pre-programmed poses and human instructions. Figure~\ref{fig:vision-teach} shows how each cycle of data collection works, where we randomly add variations to the initial gripper pose to simulate various initial object configurations. Specifically, the variations are applied via randomly sampling the grasping pose of the gripper with varying rotations along the $y$-axis and translations along the $x$- and $z$-axis. The variation for the rotation around y-axis is $\pm10^\circ$, translation around the x-axis is $\pm5\text{mm}$, and translation around the z-axis is $+5\text{mm}$ (see Figure~\ref{fig:insertion_illustration} for visualizations). Each demonstration is a triplet of $\left( P_\mathcal{O},\; P_\mathcal{S},\; T^*_{\mathcal{O}\mathcal{S}} \right)$, where \( P_\mathcal{O} \) is the initial object point cloud, \( P_\mathcal{S} \) is the scene point cloud, and \( T^*_{\mathcal{O}\mathcal{S}} \) is the ground-truth transformation that aligns the object into its target configuration. The goal object point cloud is then given by \( P^*_\mathcal{O} = T^*_{\mathcal{O}\mathcal{S}} \cdot P_\mathcal{O} \), preserving per-point correspondence between the initial and goal object configurations. These variations increase the difficulty of the task by inducing diverse viewpoints (and therefore occlusions) in the observed object point clouds, which may appear partially visible or differently oriented across demonstrations. For all three types of connectors, we collect 20 demonstrations of successful connection placements, and we split them into 16 demos for training and 4 demos for validation.

\subsection{Task Descriptions}
We designed unimodal and multi-modal placement tasks using these connectors. For the unimodal tasks, the robot inserts each of the three distinct NIST connectors into its single corresponding socket, with the primary challenge arising from significant, random perturbations to the connector's initial configuration via varying the object's initial pose in the gripper. The multi-modal task focuses on the ``Waterproof'' connector but increases scene complexity by introducing a second valid placement mode and geometrically distinct sockets as distractors. These tasks are exceptionally challenging, as it requires achieving millimeter-scale precision from partial and noisy point cloud inputs. This difficulty is compounded by the significant, randomized variations in the object's starting pose, which create diverse and occluded viewpoints that are not well supported by the sparse training samples

\subsection{Evaluation and Metrics}
To execute the predicted goal placement pose, we calculate the corresponding pre-insertion target gripper pose as an intermediate waypoint and execute the trajectory with the motion planner. We report success rates for both unimodal and multi-modal tasks, which is manually determined by observing whether the connector is inserted into the socket, with connector's pins and boundary precisely aligned with the corresponding socket. For unimodal tasks, we additionally report the translation and rotation errors with respect to the predicted gripper pose to further quantify the precision of each model's predictions and the demanding success criteria of the insertions. For all tasks, we evaluate on 20 trials with the random grasp variations shown in Figure~\ref{fig:insertion_illustration}.

\subsection{Additional Results and Analysis}

To more thoroughly investigate the precision requirements of these tasks, we further separate the trials from the unimodal insertion tasks into successes and failures, and evaluate average translation and rotation error amongst those groups. The numbers in the red parentheses \textcolor{red}{()} indicate the average error differential between failures and successes, and the dashes ``-" indicate results where no corresponding successes or failures were recorded. As Table~\ref{tab:results_table_real_additional} indicates, both Waterproof and DSUB-25 require extremely precise translation accuracy within $\sim$1 mm. Waterproof similarly requires rotation accuracy of $\sim1^\circ$, with DSUB-25 requiring $<2^\circ$ rotation error. Although SSD appears more error tolerant, it is in fact the most challenging task due to its geometry: the SSD object is relatively tall, and even small rotation errors in the point cloud prediction of its bottom face amplify into large errors in the predicted gripper pose.

%\newpage

\end{document}